\documentclass{article} 
\usepackage{arxiv,times}
\usepackage{amsmath}
\usepackage{verbatim}
\usepackage{amsfonts}

\usepackage{listings}
\usepackage{xcolor}

\colorlet{punct}{red!60!black}
\definecolor{background}{HTML}{EEEEEE}
\definecolor{delim}{RGB}{20,105,176}
\colorlet{numb}{magenta!60!black}
\newenvironment{ccr}{\fontfamily{ccr}\selectfont}{\par}
\lstdefinelanguage{json}{
    basicstyle=\normalfont\ttfamily,
    showstringspaces=false,
    breaklines=true,
    frame=lines,
    backgroundcolor=\color{background},
    literate=
     *{0}{{{\color{numb}0}}}{1}
      {1}{{{\color{numb}1}}}{1}
      {2}{{{\color{numb}2}}}{1}
      {3}{{{\color{numb}3}}}{1}
      {4}{{{\color{numb}4}}}{1}
      {5}{{{\color{numb}5}}}{1}
      {6}{{{\color{numb}6}}}{1}
      {7}{{{\color{numb}7}}}{1}
      {8}{{{\color{numb}8}}}{1}
      {9}{{{\color{numb}9}}}{1}
      {:}{{{\color{punct}{:}}}}{1}
      {,}{{{\color{punct}{,}}}}{1}
      {\{}{{{\color{delim}{\{}}}}{1}
      {\}}{{{\color{delim}{\}}}}}{1}
      {[}{{{\color{delim}{[}}}}{1}
      {]}{{{\color{delim}{]}}}}{1},
}

\newenvironment{subcolumns}[1]
 {\valign\bgroup\hsize=#1##\cr}
 {\crcr\egroup}
\newcommand{\nextsubcolumn}{\cr\noalign{\hfill}}
\newcommand{\nextsubfigure}{\vfill}

\usepackage{graphicx}
\graphicspath{ {./images/} }
\usepackage{caption}
\usepackage{subcaption}

\usepackage{hyperref}
\hypersetup{
    colorlinks=true,
    linkcolor=red,
    filecolor=magenta,      
    urlcolor=blue,
    citecolor=purple,
    pdftitle={Overleaf Example},
    pdfpagemode=FullScreen,
    }

\title{Many Episode Learning in a Modular Embodied Agent via End-to-End Interaction}

\author{\centerline{Yuxuan Sun \qquad Ethan Carlson \qquad Rebecca Qian \qquad Kavya Srinet \qquad Arthur Szlam}
\\
\\ \centerline{Meta AI Research}
}

\begin{document}

\maketitle

\begin{abstract}
In this work we give a case study of a modular embodied machine-learning (ML) powered agent that improves itself via interactions with crowd-workers.  The agent consists of a set of modules, some of which are learned, and others heuristic.  While the agent is not ``end-to-end'' in the ML sense, end-to-end interaction with humans and its environment is a vital part of the agent's learning mechanism.   We describe how the design of the agent works together with the design of multiple annotation interfaces to allow crowd-workers to assign credit to module errors from these end-to-end interactions, and to label data for an individual module.  We further show how this whole loop (including model re-training and re-deployment) can be automated.  Over multiple loops with crowd-sourced humans with no knowledge of the agent architecture, we demonstrate improvement over the agent's language understanding and visual perception modules. 
\end{abstract}

\section{Introduction}
Present day machine learning (ML) research prioritizes end-to-end learning.  Not only are end-to-end models able to achieve excellent performance on static tasks, there is a growing literature on how to adapt pre-trained networks to new tasks, and large pre-trained models can have impressive zero-shot performance on unseen tasks.  In the setting of embodied agents, this manifests as agents actualized as monolithic ML models, where inputs to the model are the agent's perceptual sensors, and the model's outputs directly control agent actions.   There are now a number of environments designed for the training of end-to-end embodied agents \cite{beattie2016deepmind, savva2019habitat, guss2019minerl, petrenko2021megaverse}, and there is hope (and some evidence) that the same sort of transfer and adaptability seen in language and vision models will carry over to the embodied agent setting.

Nevertheless, agents implemented as fully end-to-end ML models are rare in production systems (or in real-world embodied agents, a.k.a. robots).  While this in part is a symptom of the rapid improvement and scaling in the literature and the lag in technology transfer, these systems require performance and safety guarantees that are still not easily obtainable from end-to-end ML models; and must be maintainable by human engineers.    On the other hand, it is difficult for pipelined agents
to learn from experience once deployed.  Instead, human engineers design a module, collect and collate data for it, train the appropriate ML model, and then deploy it.  Thus the agent's abilities don't scale directly with the experience it receives, but rather with the amount of human engineering power that can be brought to bear in building the modules.  To somewhat oversimplify, engineers trade off ML scalability (the ability to learn new things through interaction, without engineering investment) for modularity, serviceability, and interpretability. 





This work is a case study
of automating self-improvement via interactions with people in a pipelined ML-powered agent.  The agent consists of a set of modules, some of which are learned, and others heuristic.  The agent is not ``end-to-end'' in the ML sense, but end-to-end interaction, in the sense of players interacting with the full agent system, is a vital part of the agent's learning mechanism.  

Our main result is demonstrating that subsystems in the agent can be improved with signal from these end-to-end interactions, further augmented with annotation tasks routed from marked errors.  The crucial point is that with appropriate UX (user experience) design,  sets of crowd-workers with no special training (and in particular, without any knowledge of the agent architecture or any agent modules) are able to assign credit to module errors and annotate them at scale.  We fully automate a loop of human-agent interaction, credit-assignment, module-data-annotation, model-retraining and re-deployment, and show that this successfully improves the agent's neural semantic parsing module (a finetuned BERT encoder with decoder trained from scratch) and visual perception module (a CLIP-style vision model) over multiple iterations. We thus give evidence that it is possible to keep modularity without giving up ML scalability in this setting.

\begin{figure}
    \centering
    \includegraphics[width=15cm]{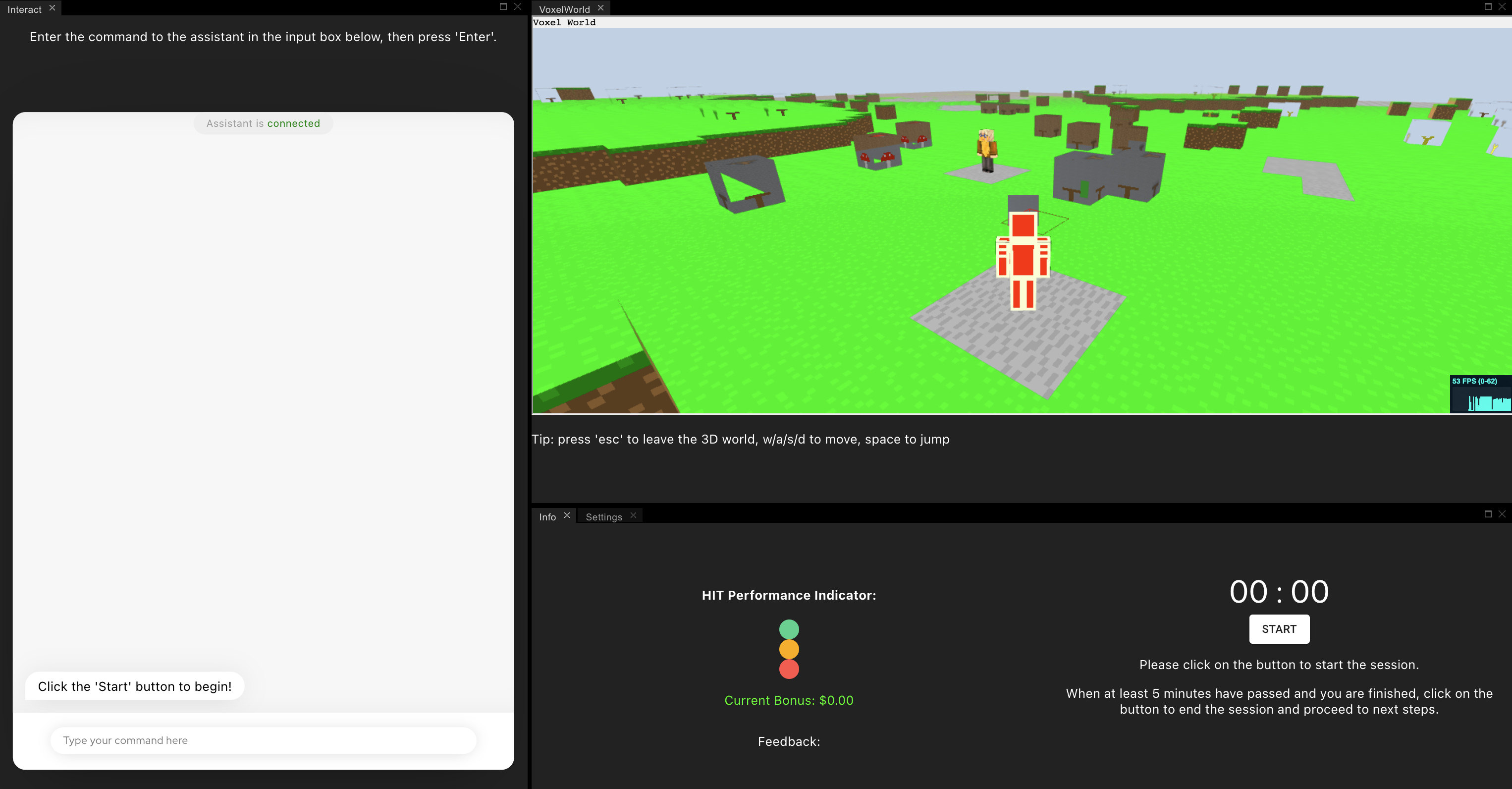}
    \caption{An image of the dashboard user interface seen by crowd workers in production.}
    \label{fig:world}
\end{figure}



\section{Setting and Methods}
\subsection{The setting}
We describe the agent architecture and the world in which it lives.

\subsubsection{World}
\label{sec:world}
The agent is embodied in a three dimensional voxel world.  Each voxel can be occupied by space or an impassable block of material.  Movement is possible in any direction, as long as the voxel is unoccupied; and the agent moves via discrete steps of size one voxel.  The agent also can turn to look in any direction; and so its pose can be represented by a $(x, y, z, \text{pitch}, \text{yaw})$ tuple.  In addition to being able to act by changing its body or head position, the agent can point at rectanguloid regions of space (by visibly flashing them), and can ``speak'' in text.  The agent can also place blocks of various colors, or destroy them. 

A human player co-occupies the world with the agent.  The human player's pose is determined by an $(x, y, z, \text{pitch}, \text{yaw})$ tuple.    The human player can also speak in text to the agent, and the agent can see the human player's pose (including the pitch and yaw, allowing it to decide what the human is looking at). The human can place blocks and destroy blocks.  See the interface in Figure \ref{fig:world}. 

\subsubsection{The agent}
We use a Droidlet agent \cite{pratik2021droidlet}.  The agent's perceptual input includes the location of the agent's self pose, the player's pose, the location and type (i.e. material) of each block in space, and the chat history.    
It is equipped with heuristic perceptual methods to recognize connected components of blocks and the local ground plane. It is also equipped with a CLIP-style visual perceptual module which can predict segmentation of objects referred with natural language descriptions. It makes use of a BERT-based semantic parsing model further described in Section \ref{sec:parser} as its natural language understanding (NLU) ``perception''.

The agent has heuristic, scripted ``Tasks'' that allow execution of atomic programs like movement to locations in space, re-orienting pose, pointing, or placing blocks.   The agent has a limited, scripted dialogue capabilities (also implemented as Tasks) to ask clarification questions to the human players when the agent is uncertain about something (for example if the human says ``destroy the red cube'' and there are two red cubes in the scene, the agent might decide to ask the human which cube they meant, by pointing at one of them).

The parameters of these Tasks are provided by a ``Controller'' module that inputs a partially specified program in the agent's domain specific language (DSL), either from the output of the NLU module or from the agent's intrinsic behaviors, and, using the agent's memory system, fully specifies the program.  See \cite{pratik2021droidlet} for more details

\subsection{NLU details}
\label{sec:parser}
The agent uses a neural semantic parser (NSP) to convert commands from players into partially specified programs in the agent's DSL; these are fully specified into executable Tasks in the agent's interpreter, see \cite{pratik2021droidlet} for details.   The neural semantic parser is one of the two ML module that has been shown to improve over the course of our experiments in this paper.

The agent's DSL is similar to the one described in \cite{srinet2020craftassist}, using the same top-level commands (Move/Dance, Build/Copy/Destroy/Dig, Stop/Resume), but the children of these have been expanded.  For example, a ``Copy'' top-level command might take a ``ReferenceObject'' (corresponding to some object in the world) as a child, and the possible queries to specify that ReferenceObject have been expanded from \cite{srinet2020craftassist}.  The full grammar is included in the supplemental, and some examples are displayed in Figure \ref{fig:lf_examples}.

The architecture of agent's semantic parsing model is similar to the one described in \cite{srinet2020craftassist}.  It is an encoder-decoder seq2seq model where the encoder is finetuned from  BERT \cite{Devlin2019BERTPO} using \cite{huggingFace}, and the decoder is trained from scratch.  In order to use a sequence based decoder, we linearize the target logical forms in depth-first order. 

\subsection{Vision module details}
\label{sec:parser}

The agent uses a 3D referring object segmentation model to detect objects specified by ungrounded natural language expressions. The vision module takes in the 3D voxel world state, as well as text-based "ReferenceObject" descriptions extracted from DSL of NLU outputs and output 3D segmentation masks. The vision model is the second ML module that has been shown to improve over the course of our experiments in this paper.

The architecture of agent's vision model is similar to \cite{radford2021learning}, but in a 3D setting. It consists of a Voxel Encoder which is 4-layer convolutional neural network trained from scratch, a Text Encoder which we directly took from CLIP \cite{radford2021learning}. The output of the model is the probability of being classified as the referred object on each voxel. All voxels with probability high than a threshold are then used to construct the segmentation mask. See more details of model architecture in Appendix \ref{sec:vision_module_detail}.

The initial dataset used to train the baseline vision model is generated using rule-based scripts. More specifically, we construct schematics of various shape objects using mathematical formulas in a flat 3D voxel world described in \ref{sec:world}. Full details and some examples are included in Appendix \ref{sec:vision_module_detail}.

\begin{figure}[h!]
\fontsize{7pt}{8pt}\selectfont
\begin{subcolumns}{0.48\columnwidth}
\begin{subfigure}{0.48\columnwidth}
\vspace{.5cm}
  \centering
"dig a moat around the fort":
\begin{lstlisting}[language=json]
 "action_sequence": [
  {"action_type": "DIG", 
   "location": {
     "relative_direction": "AROUND", 
     "reference_object": {
       "filters": {
         "where_clause": {
           "AND": [{"pred_text": "has_name", 
                    "obj_text": [0, [5, 5]]}]}
                    }}},
    "schematic": {
      "filters": {
        "where_clause": {
          "AND": [{"pred_text": "has_name", 
                   "obj_text": [0, [2, 2]]}]
          }}}}]
\end{lstlisting}
\end{subfigure}
\nextsubcolumn
\begin{subfigure}{0.48\columnwidth}
\centering
"move to the left of the cube": 
\begin{ccr}
\begin{lstlisting}[language=json]
"action_sequence": [
  {"action_type": "MOVE",
   "location": {
      "relative_direction": "LEFT",
      "reference_object": {
        "filters": {
          "where_clause": {
            "AND": [{"pred_text": "has_name", 
                     "obj_text": [0, [6, 6]]}]
            }}}}}]
\end{lstlisting}
\end{ccr}  
\end{subfigure}
\nextsubfigure
\begin{subfigure}{0.48\columnwidth}
\centering
"build a box": 
\begin{ccr}
\begin{lstlisting}[language=json]
 "action_sequence": [
  {"action_type": "BUILD", 
   "schematic": {
     "filters": {
       "where_clause": {
         "AND": [{"pred_text": "has_name", 
                  "obj_text": [0, [2, 2]]
                }]
         }}}}]
\end{lstlisting}
\end{ccr}
\end{subfigure}
\end{subcolumns}
\caption{Some examples of commands that can be parsed in the agent's DSL.  Fields of the form $[x, [y, z]]$ where $x$, $y$, and $z$ are numbers are {\it spans} of text (e.g. the $y$ through $z$th tokens on the $x$th text input).  Fields with keys "filters" correspond to queries to the agent's database. \label{fig:lf_examples}}
\end{figure}

\subsection{Learning from Humans}
Human workers are connected with an agent online. They interact with it through their web browser, where we render the agent and a representation of the world. Interaction data is gathered through crowd-sourced tasks where the workers are instructed to issue free form commands to the agents, using a category of actions from a suggested list of agent's capabilities (eg. 'build', 'destroy'). The workers are given no other instructions about what type of commands to give, other than to be creative and diverse.

After each command, workers are prompted as to whether the task was carried out correctly end-to-end, whether the command was correctly understood, and whether the agent correctly perceived the objects that the workers referred to. If the player marks that the command was not correctly understood, then this command and the agent's parse are recorded as an NLU error. If the player marks that the agent did not correctly perceived the objects that the workers referred to, then this command, along with a snapshot of the world state, are recorded as a vision error.

\begin{figure}[ht]
    \centering
    \includegraphics[width=16cm]{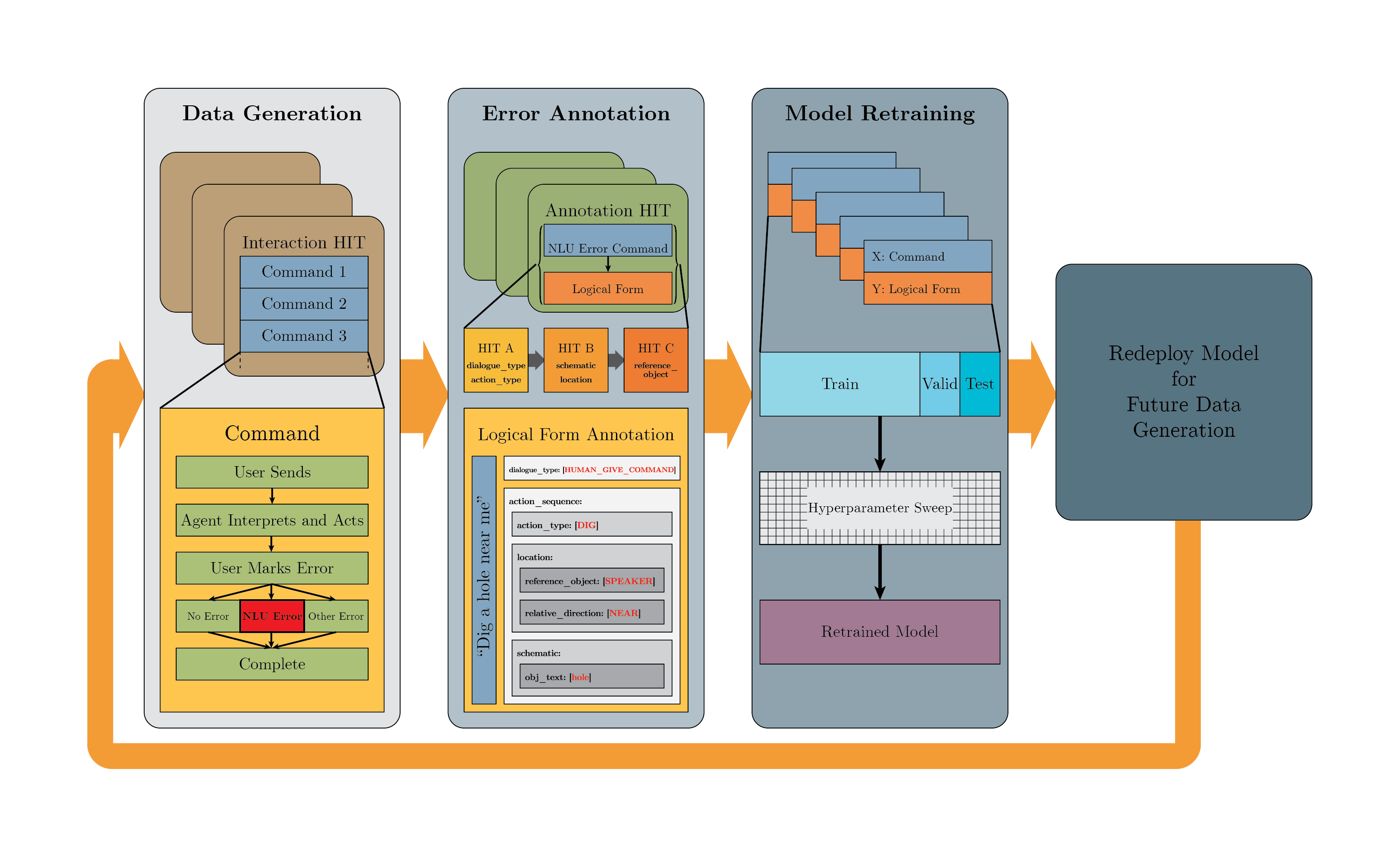}
    \caption{A diagram showing the Droidlet agent's NLU module improvement with crowd-worker interactions.  Note that the workers have no knowledge of the agent architecture. The logical form above has been simplified for clarity.}
    \label{fig:hitl}
\end{figure}

These marked NLU errors are then routed to {\it another} set of qualified crowd workers who write the ground truth parses for these commands.  These parse annotation tasks are further distributed into small tasks consisting of 1-3 questions that determine the annotation of a particular node in the parse tree.  These full annotations are then used as training data to improve the NLU model offline. The same applies to the vision errors where they are routed to a vision annotation tool for annotation, and then used as training data to improve vision models offline. These retrained models are then re-deployed before the next set of human interactions with the agent. Figure \ref{fig:hitl} has a diagram showing the agent learning pipeline.  The entire pipeline operates autonomously, from launching interaction jobs, to error annotation, to model retraining, to re-deployment.
\subsubsection{Challenges of crowd-sourcing Human-Agent Interactions}

Working with humans in the loop involves challenges that go beyond model architectures and learning algorithms.  Apart from making tools that are effective for cooperative people, it is necessary to plan for annotators that will sometimes behave erratically, or even adversarially.  

A major issue (common to many crowd-worker deployments) has been - dealing with workers who, covertly or overtly, try to cheat their way through the task.  Cheating in this case could mean not doing the task at all, trying to game our qualification criteria, or simply doing the bare minimum to pass but not engaging with the task.  The combination of the following methods has allowed for very high quality data collection:
\begin{itemize}
\item Workers must first qualify for our interaction task by answering a simple set of questions to prove they are not a bot and are capable of reading the instructions.
\item We disable the submission button until a basic list of criteria have been met, and we don't advertise what those criteria are beyond the task instructions.
\item By offering performance incentives, we make it more profitable to not cheat than to cheat.
\item We blacklist workers who repeatedly perform poorly.
\item We ask workers to reflect on their own performance, which facilitates perspective-taking and improved performance on repeated iterations of the task. \cite{10.1145/2145204.2145355}
\end{itemize}

Even workers who are not acting adversarially can present challenges to development.  They may not understand the instructions if not presented clearly, their knowledge of the English language may vary, and they may not have a strong aptitude with technology to navigate the interface.  These constraints necessitate a focus on usability and user testing throughout the life cycle of the project.

While the human factor presents varied challenges to the development of the agent interface, it has also created a continuous feedback cycle that facilitates overall agent improvement.  We have many users of the system issuing thousands of commands, some of which cause the agent to crash or behave in unexpected ways.  We would not discover these edge cases very quickly on our own.

\subsubsection{Error Routing}
\label{routing}
There could be several types of issues that cause the agent to fail to execute a command.  One example is that the user asks the agent to do something it that is not expressible in its DSL (``let's play chess'') or that is in its DSL, but part of the command refers to something the agent does not know (``build a camel'' and the agent does not know what a ``camel'' is).  The agent can also fail because its visual perception module did not recognize an object, because it did not correctly retrieve the right information from memory, and the focus of this paper: because the NLU model failed to accurately parse the command.

Differentiating between these types of failure is essential for being able to route the correct data to the correct annotator.  In the current operation, only commands that are marked as containing an NLU error are sent to be annotated to so that ground truth for those commands is added to the training dataset.

This process of differentiating between types of errors is executed using a decision tree that is presented to the worker one question at a time.  In the Appendix there are examples of this decision tree in Figure \ref{fig:nlu_error}, which represents the correct error marking flow after an NLU error, and Figure \ref{fig:task_error}, which represents the error marking flow after a non-NLU task error.

\section{Related Work}
There is a large literature on human in the loop machine learning, see \cite{hitl_survey} for a survey.

Our setting is an embodied agent with a language interface. There is existing work showing improvement after multiple rounds of re-deployment with dialogue agents, for example \cite{hancock2019learning, shuster2021dialogue, kiela2021dynabench}.

Some prior work building towards sophisticated interactive tools for ``machine teaching'' \cite{simard2017machine}, where ML naive users are able to guide model training towards high accuracy and coverage, has been considered in the literature and many such tools exist as deployed services, for example \cite{ratner2017snorkel} (commercialized at \url{https://snorkel.ai/}) or \url{https://scale.com/}.  These are superior to the re-deployment loop described in this work in the sense that the model re-training occurs ``in-session'', and the machine teacher can immediately see the results of their annotations and adjust accordingly.  Furthermore, these also have tools for automatically generating labeled data from rules or automating data augmentation.  However, the work described in this case study is complementary to these, in that it focuses on automating the end-to-end data-collection and retraining of ML models that are important internal components of an embodied agent.  We give evidence that fully modular ML systems will be able to self-improve even if gradients cannot pass from one part of the system to another.  In future work, we hope to combine our system with responsive in-session learning as described in these services. 

Our work is inspired by \cite{wang2017naturalizing}, where multiple rounds of users build up a semantic parser for a voxel world editor.  In \cite{shah2021minerl} the authors propose a competition to train embodied agents in a voxel world through language descriptions.  Our work is also related to \cite{suhr2019executing}, where the authors build an interactive environment where (embodied) players and agents (playing as the role of a language issuing ``leader'' with full observability or faster moving ``follower'' with partial observability) collaborate via natural language to collect cards by moving to their spatial locations.  A followup \cite{kojima2021continual} is especially relevant; in that work they show how multiple rounds of learning can continue to improve the language generation capabilities of a ``leader'' model.  In addition to the embodied agents and players, our work shares with 
 \cite{kojima2021continual} multiple rounds of data collection and the use of player feedback after ``execution'' to label examples.   However, the key difference is that in \cite{kojima2021continual} the agent is a single ML model, whereas in this work, we aim to show that credit can be assigned to different components in a modular system, the data for the component can be annotated, and the component re-trained without any engineer intervention.

There are several works showing how humans can interactively teach robotic agents, for example \cite{saxena2014robobrain,  paxton2017costar, mandlekar2018roboturk, Cabi2019a, mandlekar2020human}. In \cite{saxena2014robobrain}, the authors demonstrate large-scale crowd-sourcing of data for perceptual and knowledge-base components of a robotics system. In \cite{mandlekar2018roboturk, mandlekar2020human} crowd-workers are connected with robotic manipulators to demonstrate movements or parts of movements.  COSTAR \cite{paxton2017costar} is a modular system for teaching robots to carry out tasks using behavior trees. Our work is similar to COSTAR in that it is built on a modular system with perception decoupled from action generation; but in this work we focus on the infrastructure for crowd-sourcing annotations, rather than mechanisms for live human teaching.

Finally, our work builds on the ideas of \cite{carlson2010toward,mitchell2018never}.  Our hope is to demonstrate progress towards {\it embodied} incarnations of these.  We show that the system is able to improve even when the module being supervised is far from the end-to-end experience the agent receives, and that architecture-naive crowd-workers operating as group  can route errors to appropriate modules and teach them at scale.

\section{Results}

\subsection{NLU Error Collection}

In Table \ref{tab:errorfunnel} the results of the NLU error generation funnel are reported.  In total, over the course of the experiments, we collected $18,163$ de-duplicated commands.  


In early runs, we found that training only on new data where the NLU model failed led to feedback effects.  We updated our protocol to re-train using {\it all} de-duplicated commands at each iteration (including the ones the model correctly parsed).  We leave methods for balancing the cost of labeling against distributional stability for future work. 

Even though we annotated all of the commands on later re-deployments, we calculated the accuracies of workers in routing errors - in ongoing and future work we expect to have several active ML models across different modules of the agent.  
Workers are relatively precise: 89\% of the time they mark a command as resulting in an NLU error, it turns out to be a true error.  However,  we estimate only 43\% of NLU errors are marked.  This is an estimate because it can only be calculated for commands for which there has ever been an annotation.  Future work on interaction task design will attempt to address this discrepancy.


\subsection{Vision Module Error Collection}

We started to collect vision module errors from 11$_{th}$ iteration. After NLU module demonstrated promising results in this setting in the first 10 iterations, we integrated vision modality in to demonstrate heterogeneous learning capabilities. Over the course of following 12 iterations, we collected 695 marked vision errors, and 469 of them are successfully annotated and added to the initial dataset of size 1538.

\begin{table}
    \begin{center}
    \begin{tabular}{||l c||} 
     \hline
     \textbf{Pipeline Stage} & \textbf{Number of Commands} \\
     \hline\hline
     All Commands & 22,685 \\ 
     \hline
     De-duplicated, Valid Commands & 18,163 \\
     \hline
     Marked Agent Errors & 7,461 \\
     \hline
     Marked NLU Errors & 2,559 \\
     \hline
     Marked NLU Errors Successfully Annotated & 2,403 \\
     \hline
     Marked "True" NLU Errors & 2,138 \\
     \hline
     \textit{All Known NLU Errors} & \textit{4,944} \\
     \hline
    \end{tabular}
    \end{center}
    \caption{The number of commands for which each row description applies.  "All Commands" refers to all commands from the data presented here.  "De-duplicated, Valid Commands" refers to the subset of 'All Commands' that are unique and ask the agent to perform a task within its capabilities.  "Marked Agent Errors" refers to the number of times a worker indicated, after issuing a command and observing the resulting agent behavior, that the agent failed to perform the task.  "Marked NLU Errors" refers to the subset of 'Marked Agent Errors' for which workers indicated the agent did not understand the command, based on a report of the NSP output.  "Marked NLU Errors Successfully Annotated" refers to the subset of 'Marked NLU Errors' for which a ground truth logical form was successfully added to the data set through the annotation process.  The remainder were outstanding at the time of model retraining and redeployment but remain accessible for later use.  "Marked 'True' NLU Errors" refers to the subset of 'Marked NLU Errors Successfully Annotated' for which the ground truth annotation varied from the NSP inference.  The ratio of the two previous values forms the worker error marking precision.  "All Known NLU Errors" refers to the subset of 'De-duplicated, Valid Commands' for which a) a ground truth logical form exists in the data set and b) the ground truth annotation varied from the NSP inference.  The ratio of 'Marked "True" NLU Errors' to 'All Known NLU Errors' forms the estimate of worker error marking recall.}
    \label{tab:errorfunnel}
\end{table}

\subsection{NLU \& Vision model improvements}
\label{sec:nlu_improve}
For NLU models, We ran 22 iterations of the full interaction $\rightarrow$ routing $\rightarrow$ annotation $\rightarrow$ retrain pipeline.  The first 5 iterations were run over a period of 3 weeks.  In these,  we did not re-deploy the NLU model after an iteration.  For the next 18 iterations, we redeployed the re-trained model over each iteration. 

For vision models, the same baseline model is deployed for the first 10 iterations. Then we started to collect vision errors, annotate errors, retrain and re-deploy vision models starting from 11$_{th}$ iteration.

In order to measure model improvements, for each new tranche of data from the iterations, we randomly split it into a train, valid and test set. We then build a sequence of training data sets R$_n$ which are the union of the first $n$ training sets, V$_n$ which are the union of the first $n$ validation sets, and T$_n$, the union of the first $n$ test sets.  For NLU module, R$_0$ is taken from \cite{srinet2020craftassist} and for vision module, R$_0$ is templated data we generated using rule-based scripts (see more details in Appendix). Both of them are used to train the initial deployed NLU \& vision model, respectively.

For each tranche of data $n$, we compare three models.  The first is the baseline, trained on R$_0$.  The next is the episode-updated model, trained on R$_n$ (trained the same way as the model that was used for obtaining R$_{n+1}$).  Finally, we take the episode-updated model trained on R$_n$, and then finetune it on R$_0$; we call this the ``re-biased'' model.

\begin{figure}
\centering
  \includegraphics[width=82mm]{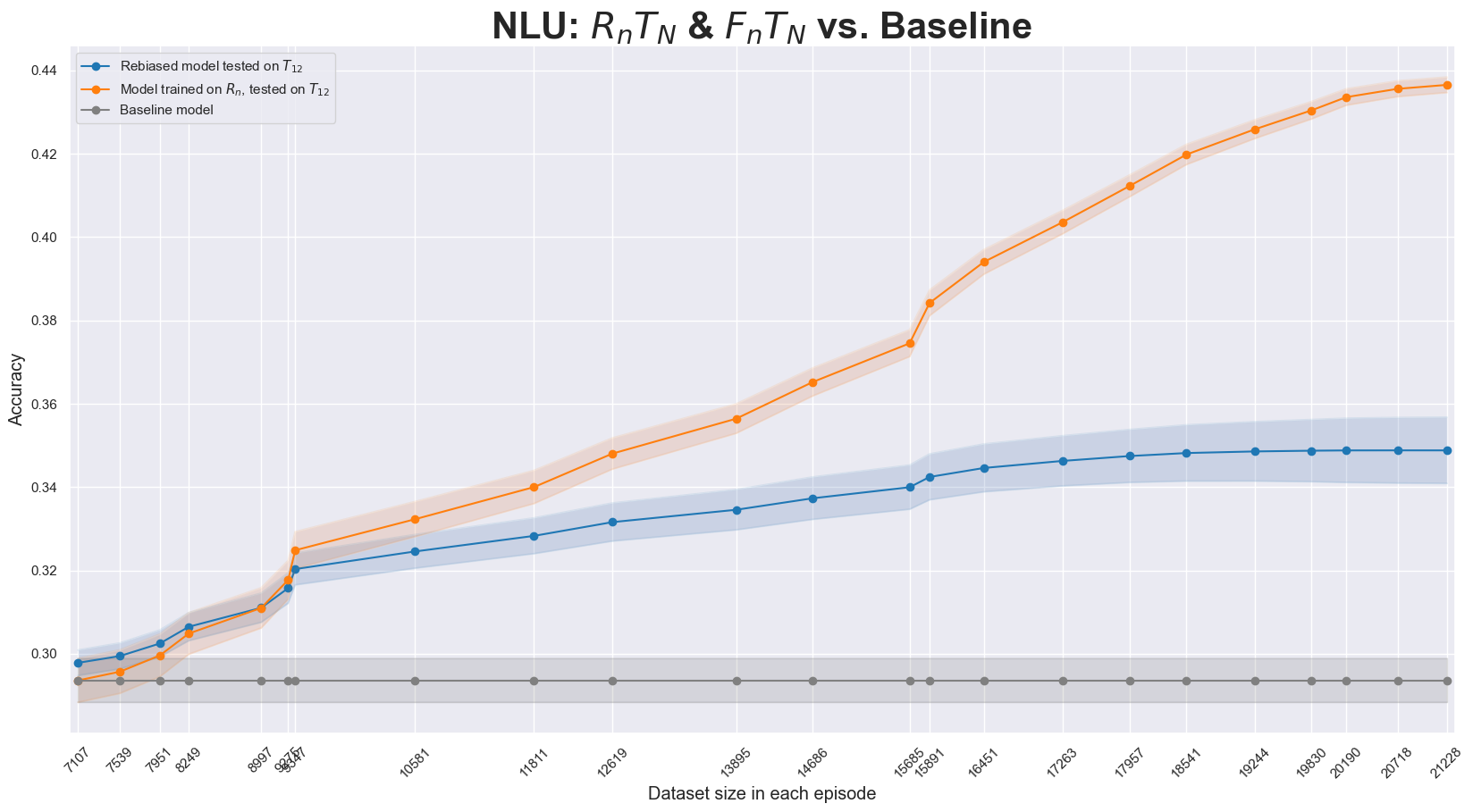}
  \includegraphics[width=82mm]{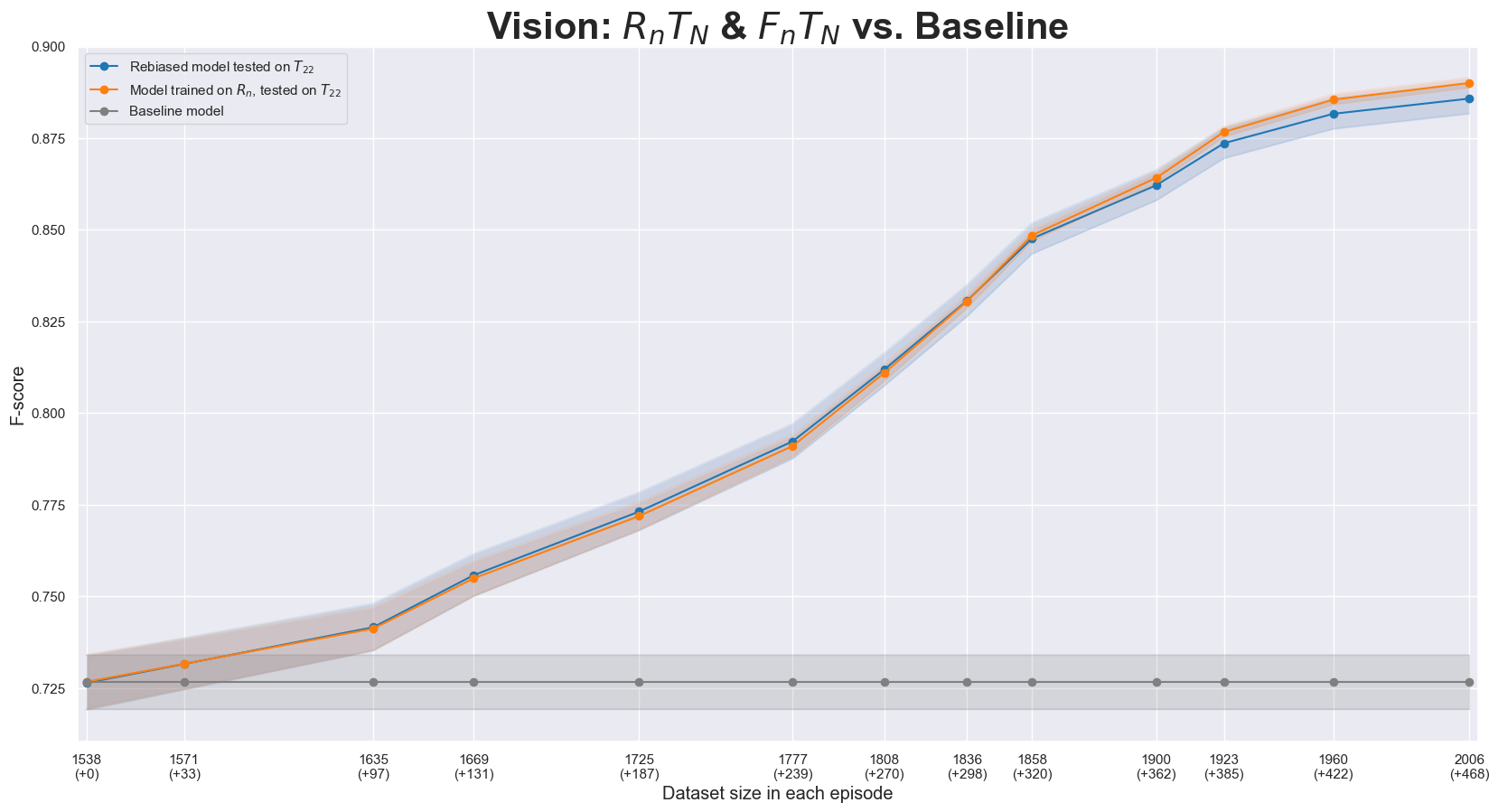}
\caption{Episode-updated model tested on T$_{all}$ (union of all collected test data). We can see a steady improvement on the final test set in the episode-updated models over the baseline model. Orange is continually learned, Blue is re-biased, and Gray is baseline.}
\label{fig:FnVN}

\centering
  \includegraphics[width=82mm]{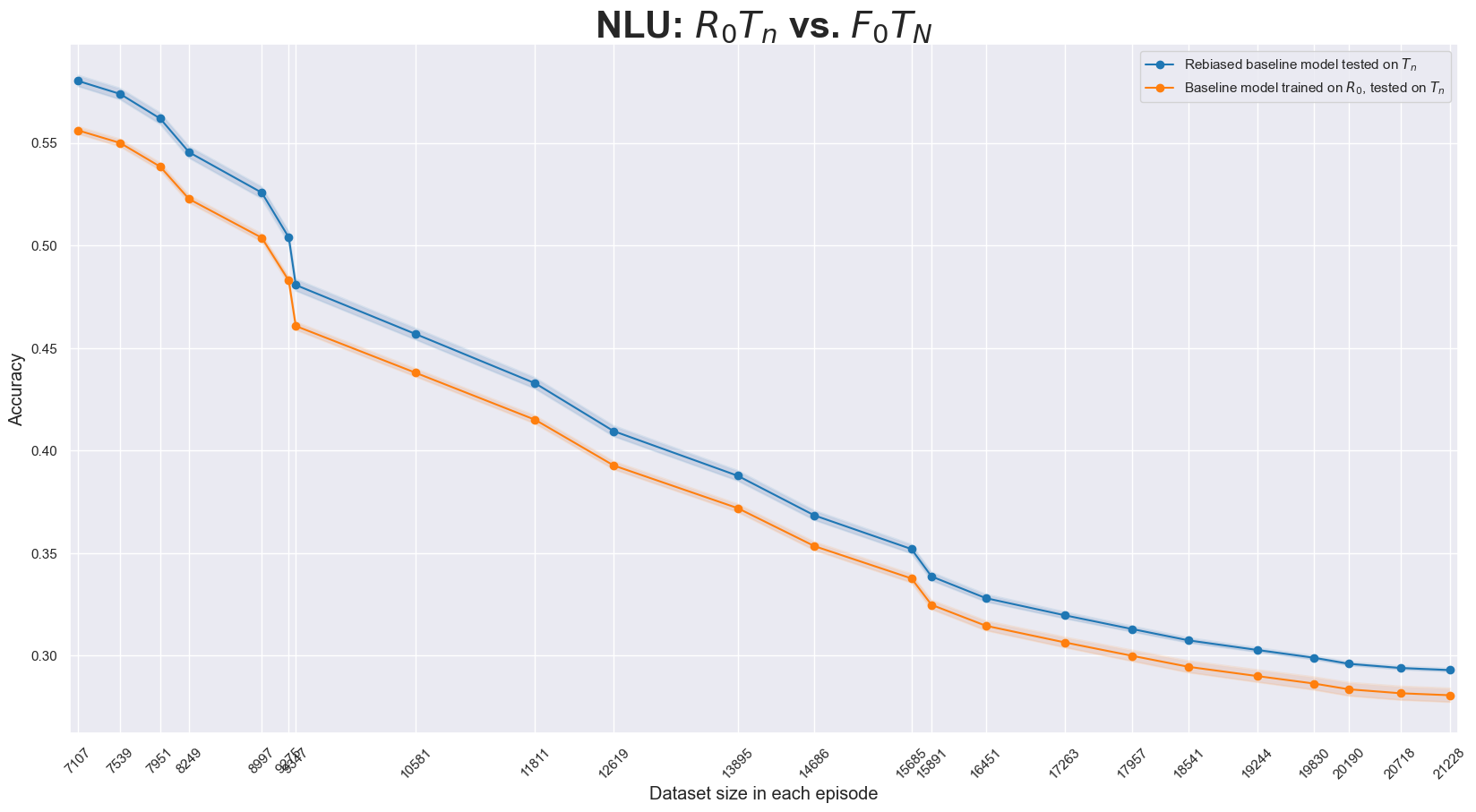}
  \includegraphics[width=82mm]{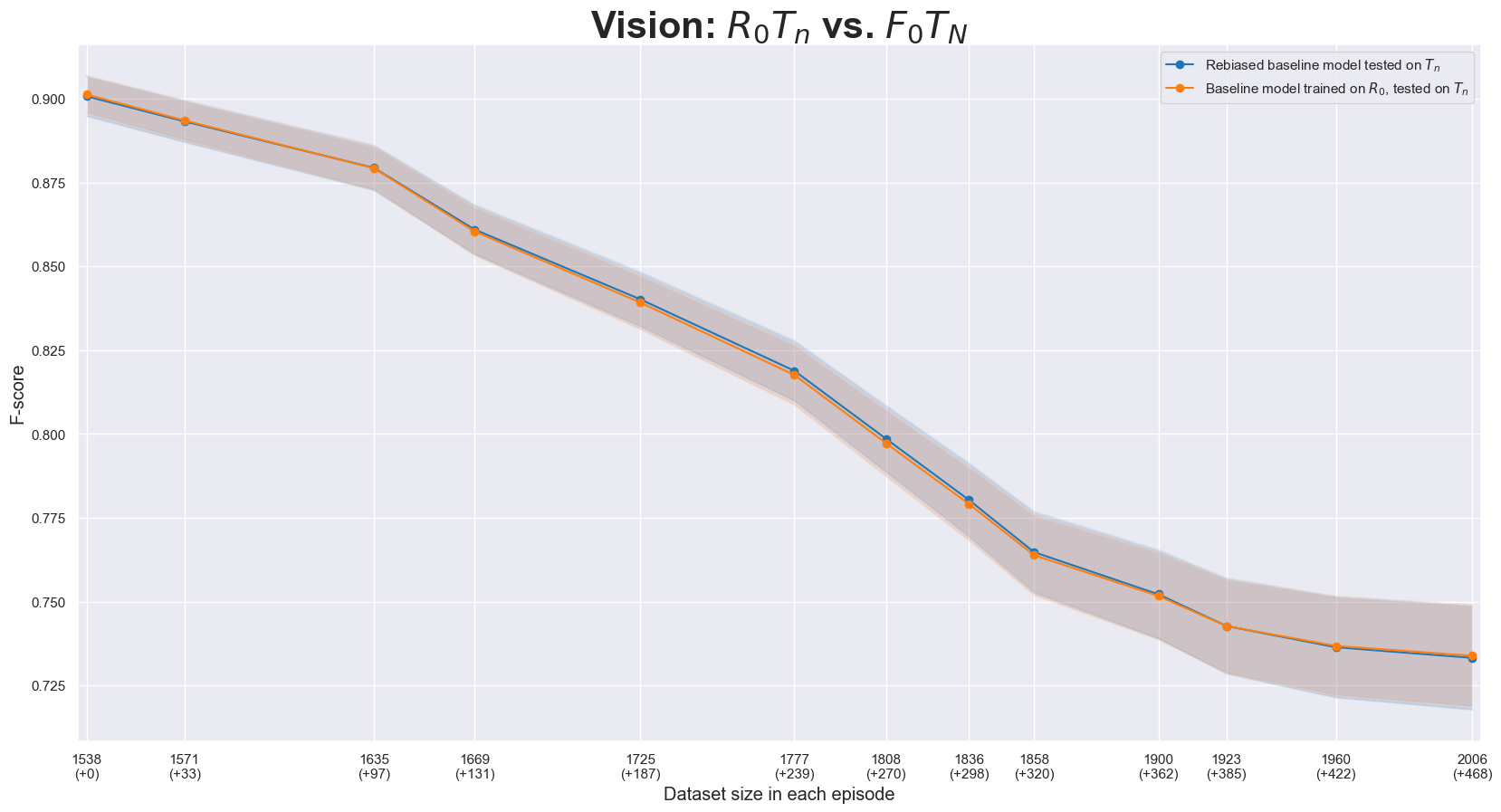}
\caption{Baseline model tested on T$_{n}$ (union of test data over tranches). The performance decay indicates the tasks become harder over tranches (i.e. unseen data points are added to dataset continually)}
\label{fig:F0Vn}

\centering
  \includegraphics[width=82mm]{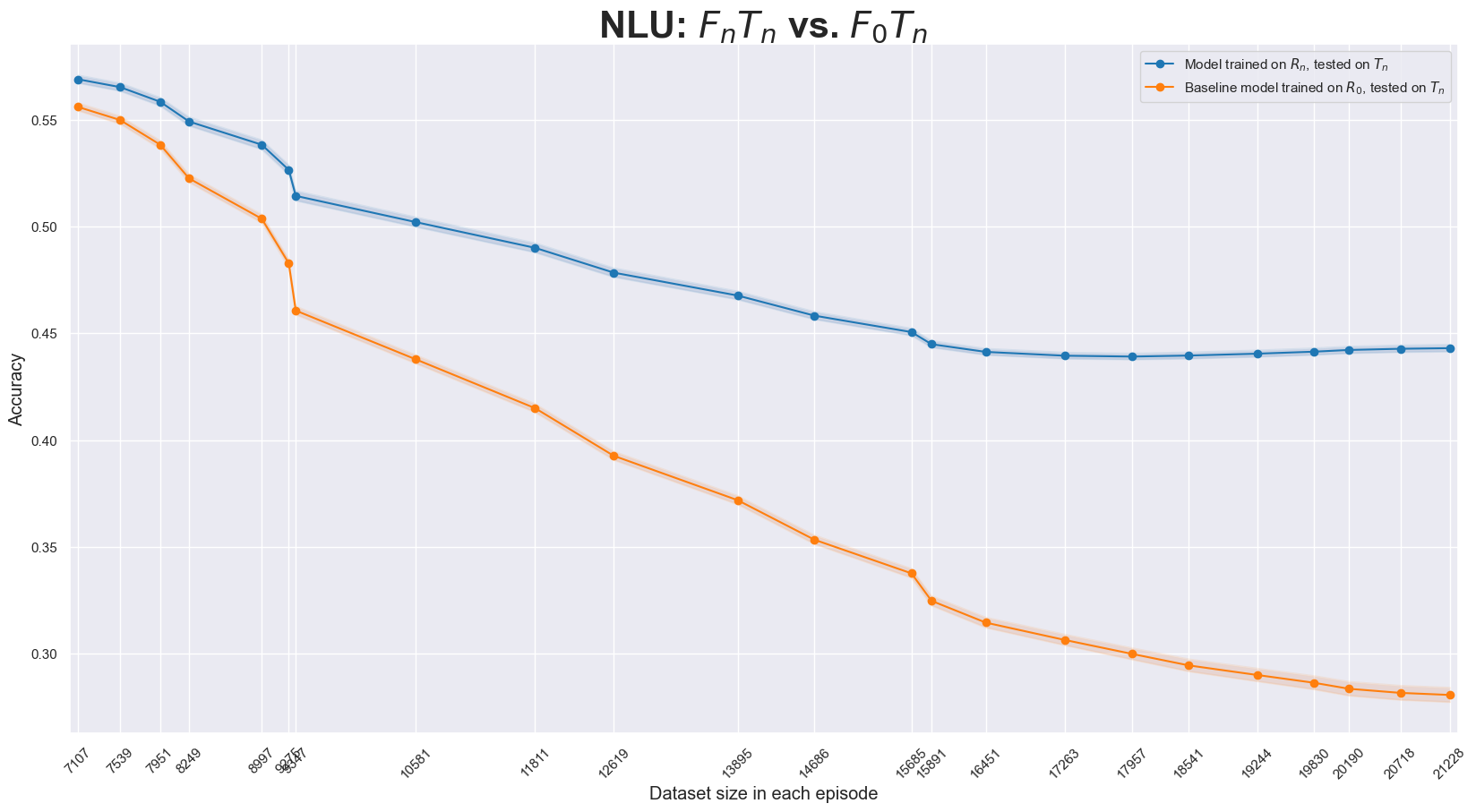}
  \includegraphics[width=82mm]{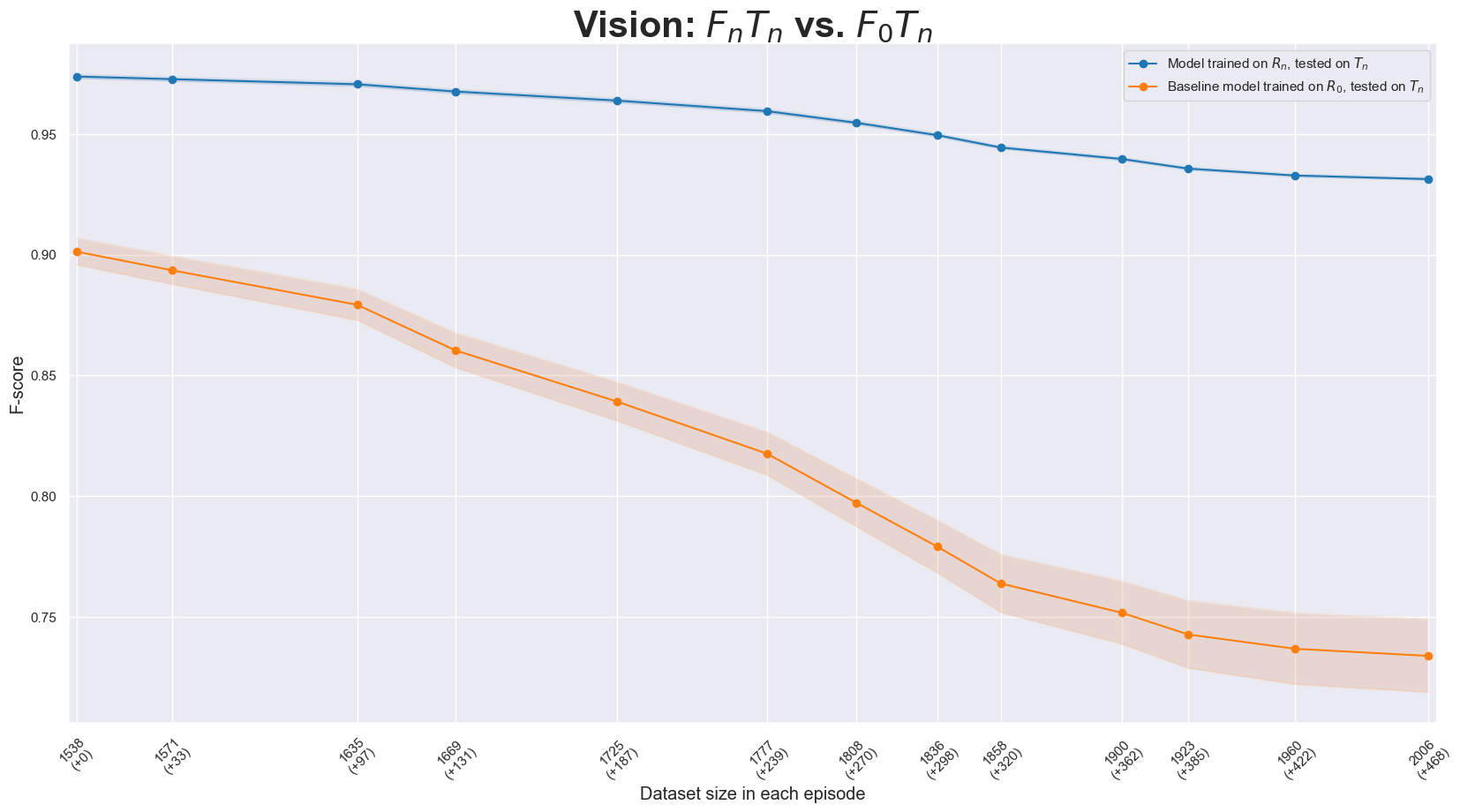}
\caption{Episode-updated model tested on T$_{n}$ (union of test data over tranches) vs. Baseline model performance tested on T$_{n}$. Episode-update models outperform baseline models consistently over each tranche.}
\label{fig:Fn_Vn}

\centering
  \includegraphics[width=82mm]{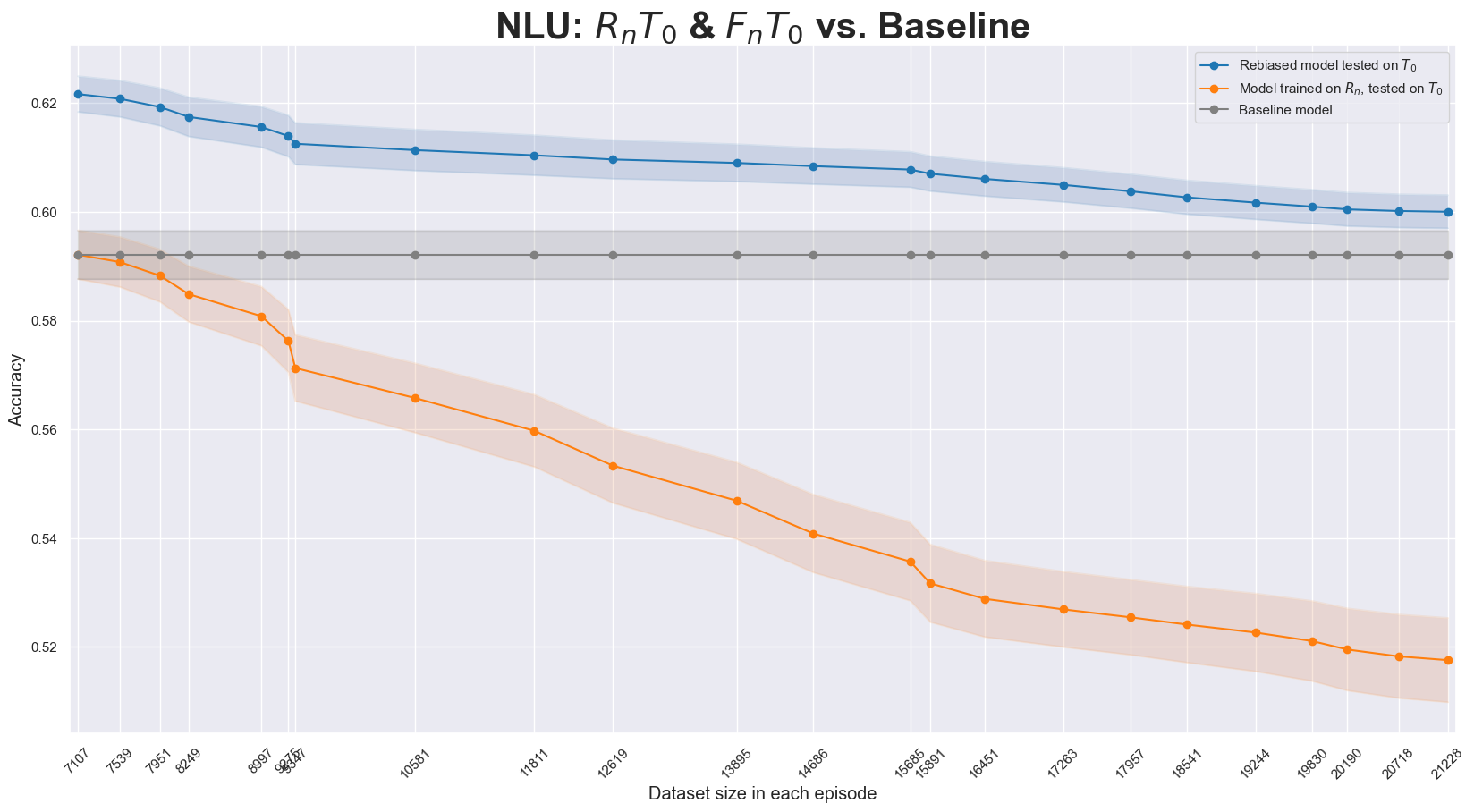}
  \includegraphics[width=82mm]{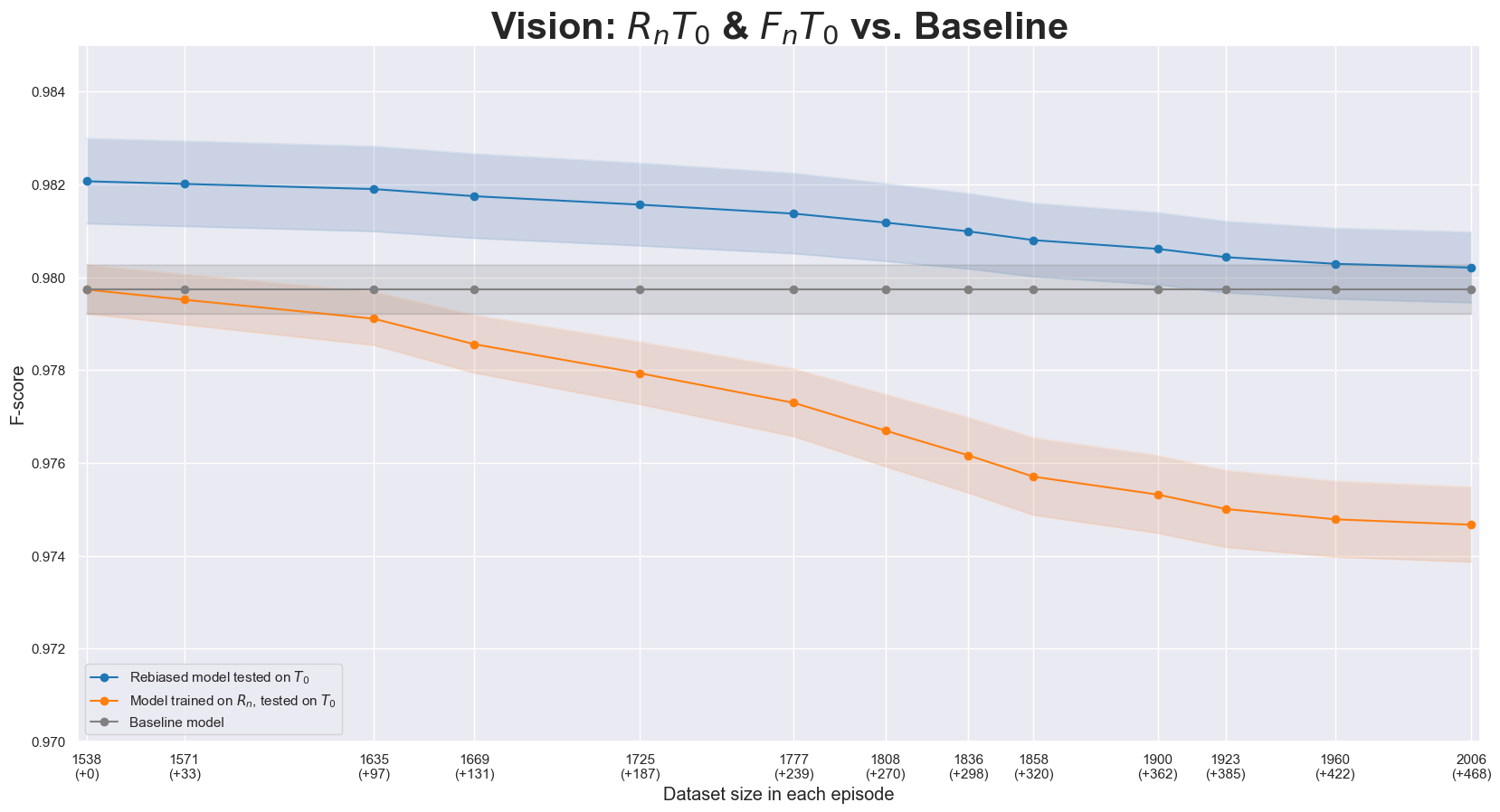}
\caption{Episode-updated model (Orange) tested on T$_{0}$ (the initial test data). Distribution shift caused the performance decay, and model re-biasing (Blue) mitigated this problem with some performance improvement over baseline (Gray).}
\label{fig:FnV0}
\end{figure}

We repeated the model training 5 times for each tranche with different random seeds.  Our main results are Figure \ref{fig:FnVN}, \ref{fig:F0Vn}, \ref{fig:Fn_Vn}, \ref{fig:FnV0}. The colored lines represent mean values of model accuracy across all 5 experiments and the shaded error bands represent the standard error.

In Figure \ref{fig:FnVN}, we show the performance of models trained on R$_n$ (all the data up to the n$_{th}$ iteration) vs. the original baseline, all tested on the final test data T$_{22}$ (the union of the test sets from all tranches).  The $x$ axis is the total number of training examples used for that model, arranged in the sequence they were obtained, and the $y$ axis is model performance. We can see a steady improvement on the final test set in the episode-updated models over the baseline.  The re-biased model also improves, although not quite as much. 

In Figure \ref{fig:F0Vn}, we show the performance of baseline model trained on R$_0$ and tested on T$_{n}$ (the union of the test sets over tranches). We can see a steady performance decay on the test set of each tranche in the baseline model, indicating that the tasks are getting harder over episodes, i.e. unseen data points are added to the dataset continually.

In Figure \ref{fig:Fn_Vn}, we show the performance of models trained on R$_n$ (all the data up to the n$_{th}$ iteration) and the performance of the baseline model trained on R$_0$, both tested on T$_{n}$ (the union of the test sets over tranches). We can see that models trained on R$_n$ outperform the baseline model consistently because these episode-udpated models are trained with additional data points whose distribution is closer to the new data points in test set while baseline model is trained with only initial training dataset.

In Figure \ref{fig:FnV0}, we show the performance of models trained on R$_n$ and testing on T$_0$ (the initial test data). The episode-updated models perform worse on T$_0$ even though they are trained with a larger amount of data; but this is not surprising, as the collection procedure for the base data R$_0$ was different than R$_i$ for $i>0$, and hence the distribution is different. Specifically, for NLU module, most of the commands in R$_0$ were collected by asking crowd-workers what they might ask an agent to do offline (no interaction with the agent) and for vision module, all the objects in R$_0$ were generated using rule-based scripts; whereas in this work, the crowd workers are actually connected to the agent in a session, and interact with it, giving multiple commands in each session and able to create/destroy/modify objects in the world freely.

The re-biased model manages to keep its performance almost at the level of the baseline (while improving on the new data).  Thus model re-biasing can mitigate the performance degradation on the original data caused by the distribution difference between newly collected data and original data, while still improving on the baseline on new data.

\subsection{Annotator Experience Improvements}
 The NLU model improvement rate is a function of the quantity and diversity of the NLU system errors, and is constrained to the first order by the resources available to fund interactions with the agent. Therefore, the goal of our UI/UX (user interface and user experience) research is to efficiently generate and correctly mark as many high quality errors as possible in each interaction, and a focus on interface usability is critical to this end. We have been guided by standard usability heuristics, the most impactful of which are listed here below.

\begin{itemize}
    \item \textbf{Aligning With Design Standards} - Utilizing UI components and affordances that match user expectations, as well as reducing overall visual clutter helps reduce cognitive load of using the interface.
    \item \textbf{Forced Choices} - Providing clear, blocking choices for important UI tasks rather than relying on the user to recognize a branch in workflow and select the appropriate option.
    \item \textbf{Visual Feedback} - Implementing clear and easy to understand visual indicators of agent status as well as the quality of the interaction (number and diversity of commands) helps the workers understand our expectations better.
    \item \textbf{Performance Incentives} - Shifting to paying workers a lower base rate with incentives for good performance both: lowers the cost of data collection on a per-error basis and results in higher worker pay.
\end{itemize}

\begin{figure}
    \centering
    \includegraphics[width=4cm]{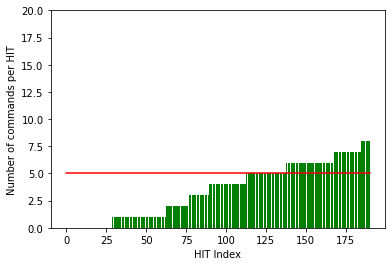}
    \includegraphics[width=4cm]{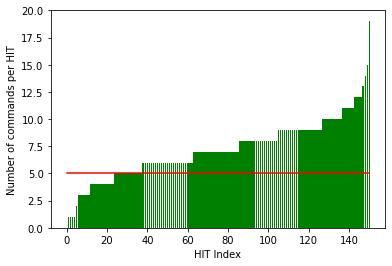}
    \includegraphics[width=4cm]{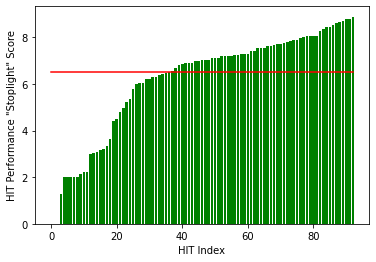}
    \includegraphics[width=4cm]{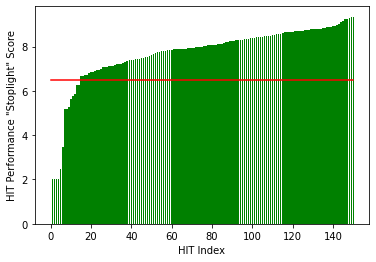}
    \caption{The charts above show the number of commands workers issued per interaction task (left two charts) and the stoplight performance score described in Section \ref{exp3} (right two charts) before (first and third charts) and after (second and fourth charts) issuing performance incentives.  The red line indicates the authors' target for each metric. There are fewer data in the third chart, because a recording bug in UI/UX Experiment 3 caused half of these data to be lost.}
    \label{fig:incentives}
\end{figure}

While there is not an obvious baseline of usability for a specific interface, Figure \ref{fig:efficiency} shows the cost efficiency improvements for each iteration as the project progressed, providing a strong validation of effort spent improving task usability.  Over the course of the four UI/UX experiments listed, the cost of collecting a single NLU error fell by 71\%.  Below is more detail on the nature of each of the four UI/UX experiments.  The experiments are cumulative, meaning each experiment includes the changes made in the previous.

\subsubsection{Experiment 1 - Clarity and Verbosity}
The goal of the first experiment was to reduce visual clutter, verbosity, and text complexity.  \cite{hirth2020taskprefs} finds that "Incomprehensible Instructions" are the single most frustrating aspect of task design when present.  Experiment 1 reduced the number of instruction words by almost half, and paginated the remaining so workers were never reading more than a few sentences at a time.  The instructions are attached in Appendix B, Figure \ref{fig:instructions}.  After the initial read during the task, instructions are hidden but available through a drop-down mechanism to further reduce clutter on screen.  The information from the instructions most relevant to producing good interaction data is copied outside the instructions window immediately next to the interaction interface for easy reference.

\subsubsection{Experiment 2 - Visual Feedback and Forced Choices}
The purpose of the second experiment was to ensure that the worker is not confused about the agent status or what to do next.  After each command, the agent must receive and interpret it, then potentially plan and carry out an action or set of actions, as well as respond to the user if appropriate.  Experiment 2 introduced messages to report the agent status periodically.

The second change in this experiment was to make the error marking decision tree described in Section \ref{routing} and shown in Figures \ref{fig:nlu_error} and \ref{fig:task_error} in the Appendix.  If error marking is a passive call-to-action, workers may not mark effectively because they may not remember to mark erroneous commands or may be eager to move on with the task.  By forcing the worker to decide one way or the other before continuing, a much higher percentage of agent errors are captured.

\subsubsection{Experiment 3 - Align with Design Standards}
\label{exp3}
Human-Computer Interaction research has shown that familiarity with an interface reduces cognitive load, and therefore increases task accuracy and reduces task completion time.  For a review if this concept see \cite{HOLLENDER2010hci}.  Experiment 3 replaced the chat interface with a UI that closely resembles that one would find on a cell phone or the help window of a website, with the purposes of better aligning with workers' existing mental model of a chat interface.

This experiment also introduced a new component to the interface - a stoplight that serves as an feedback indicator of the overall task performance to the worker.  If the light is red the worker knows that their performance is unsatisfactory, and so forth.  The metric used to determine the stoplight color is a weighted average of the log of: the number of commands issued, the diversity of commands in-session (average word edit distance between each session command), and the average creativity of the commands (word edit distance compared to all previously issued commands).  The weights and thresholds driving the stoplight indicator were empirically tuned to align with the results a worker should obtain by engaging in good faith with the task.

\subsubsection{Experiment 4 - Performance Incentives}
The final experiment in this series was meant to operationalize the stoplight introduced in the previous section by offering performance incentives based on the "stoplight score", or the score out of 10 that determines the stoplight color.  In this experiment, workers receive a lower base pay in addition to a bonus payment after completion based on their score.  Workers can see their expected bonus reported in real time after they issue each command.

This change had several notable effects.  Firstly, nearly all of the workers achieved a score in the "green" performance band, compared to the previous experiment where approximately 2/3 did, shown in Figure \ref{fig:incentives}.  Second, while the cost per interaction task went up, and therefore worker compensation per time went up, data generation efficiency measured in NLU errors per dollar actually rose.  Furthermore, workers responded to the change positively, providing qualitative feedback that in addition to increasing their compensation, the change also improved the enjoyability and clarity of the task.  This is evidence that further task gamification and/or incentives alignment may be fruitful and mutually beneficial direction for future research.

\begin{figure}
    \begin{center}
    \begin{tabular}{||l c c||} 
     \hline
     \textbf{Experiment Name} & \textbf{Number of tasks Completed} & \textbf{Data Generation Efficiency Ratio} \\
     \hline\hline
     Baseline & 17 & 1.0 \\ 
     \hline
     Exp1 - Instruction Clarity and Verbosity & 106 & 1.1 \\
     \hline
     Exp2 - Visual Feedback and Forced Choices & 197 & 2.5 \\
     \hline
     Exp3 - Design Standards Alignment & 191 & 3.1 \\
     \hline
     Exp4 - Performance Incentives & 150 & 3.5 \\
     \hline
    \end{tabular}
    \end{center}
    \caption{Table showing the the efficiency of data collection (NLU errors collected per \$) as I/UX improvements were made, reported as a ratio between the efficiency of that experiment and the baseline value before any UI/UX improvements were made.  Data generation efficiency is computed as an average over all tasks completed in that experiment.  UI/UX experiments are cumulative, not independent (each includes the changes of the previous).}
    \label{fig:efficiency}
\end{figure}

\section{Discussion}
In this work have given an example of a pipelined ML powered embodied agent that uses end-to-end interaction as a crucial part of its learning mechanism; and demonstrated that the agent, and specifically the NLU module and Vision module of the agent, can improve over multiple rounds of re-deployment with human feedback in the loop.   This is made possible in part through good UX allowing naive crowdworkers with no knowledge of the agent architecture to route errors and complete complex annotations in an assembly-line style.

In future work, we would like to extend the approaches discussed in this work to embodied agents with learnable Task executors; or even learnable memory and Controller modules.  More generally, we think these approaches will also be valuable in the context of works like \cite{dalmia2019enforcing, veniat2020efficient} that build modular ML systems that allow automatic credit assignment. The approach presented in this paper could act as a hybrid that empowers humans to teach the system at the level of its modules while automatically assigning credit when such humans are unavailable. We believe this could be more powerful than pure end-to-end or pipelined systems.

\bibliography{hitl2022}

\begin{thebibliography}{31}
\providecommand{\natexlab}[1]{#1}
\providecommand{\url}[1]{\texttt{#1}}
\expandafter\ifx\csname urlstyle\endcsname\relax
  \providecommand{\doi}[1]{doi: #1}\else
  \providecommand{\doi}{doi: \begingroup \urlstyle{rm}\Url}\fi

\bibitem[hug()]{huggingFace}
Huggingface's transformers: State-of-the-art natural language processing.
\newblock URL \url{https://github.com/huggingface/transformers}.

\bibitem[Beattie et~al.(2016)Beattie, Leibo, Teplyashin, Ward, Wainwright,
  K{\"u}ttler, Lefrancq, Green, Vald{\'e}s, Sadik, et~al.]{beattie2016deepmind}
Charles Beattie, Joel~Z Leibo, Denis Teplyashin, Tom Ward, Marcus Wainwright,
  Heinrich K{\"u}ttler, Andrew Lefrancq, Simon Green, V{\'\i}ctor Vald{\'e}s,
  Amir Sadik, et~al.
\newblock Deepmind lab.
\newblock \emph{arXiv preprint arXiv:1612.03801}, 2016.

\bibitem[Cabi et~al.(2019)Cabi, Colmenarejo, Novikov, Konyushkova, Reed, Jeong,
  Żołna, Aytar, Budden, Vecerik, Sushkov, Barker, Scholz, Denil, de~Freitas,
  and Wang]{Cabi2019a}
Serkan Cabi, Sergio~Gómez Colmenarejo, Alexander Novikov, Ksenia Konyushkova,
  Scott Reed, Rae Jeong, Konrad Żołna, Yusuf Aytar, David Budden, Mel
  Vecerik, Oleg Sushkov, David Barker, Jonathan Scholz, Misha Denil, Nando
  de~Freitas, and Ziyu Wang.
\newblock Scaling data-driven robotics with reward sketching and batch
  reinforcement learning.
\newblock Technical report, Deepmind, 2019.

\bibitem[Carlson et~al.(2010)Carlson, Betteridge, Kisiel, Settles, Hruschka,
  and Mitchell]{carlson2010toward}
Andrew Carlson, Justin Betteridge, Bryan Kisiel, Burr Settles, Estevam~R
  Hruschka, and Tom~M Mitchell.
\newblock Toward an architecture for never-ending language learning.
\newblock In \emph{Twenty-Fourth AAAI conference on artificial intelligence},
  2010.

\bibitem[Dalmia et~al.(2019)Dalmia, Mohamed, Lewis, Metze, and
  Zettlemoyer]{dalmia2019enforcing}
Siddharth Dalmia, Abdelrahman Mohamed, Mike Lewis, Florian Metze, and Luke
  Zettlemoyer.
\newblock Enforcing encoder-decoder modularity in sequence-to-sequence models.
\newblock \emph{arXiv preprint arXiv:1911.03782}, 2019.

\bibitem[Devlin et~al.(2019)Devlin, Chang, Lee, and
  Toutanova]{Devlin2019BERTPO}
Jacob Devlin, Ming-Wei Chang, Kenton Lee, and Kristina Toutanova.
\newblock Bert: Pre-training of deep bidirectional transformers for language
  understanding.
\newblock \emph{ArXiv}, abs/1810.04805, 2019.

\bibitem[Dow et~al.(2012)Dow, Kulkarni, Klemmer, and
  Hartmann]{10.1145/2145204.2145355}
Steven Dow, Anand Kulkarni, Scott Klemmer, and Bj\"{o}rn Hartmann.
\newblock Shepherding the crowd yields better work.
\newblock In \emph{Proceedings of the ACM 2012 Conference on Computer Supported
  Cooperative Work}, CSCW '12, pp.\  1013–1022, New York, NY, USA, 2012.
  Association for Computing Machinery.
\newblock ISBN 9781450310864.
\newblock \doi{10.1145/2145204.2145355}.

\bibitem[Guss et~al.(2019)Guss, Houghton, Topin, Wang, Codel, Veloso, and
  Salakhutdinov]{guss2019minerl}
William~H Guss, Brandon Houghton, Nicholay Topin, Phillip Wang, Cayden Codel,
  Manuela Veloso, and Ruslan Salakhutdinov.
\newblock Minerl: A large-scale dataset of minecraft demonstrations.
\newblock \emph{arXiv preprint arXiv:1907.13440}, 2019.

\bibitem[Hancock et~al.(2019)Hancock, Bordes, Mazare, and
  Weston]{hancock2019learning}
Braden Hancock, Antoine Bordes, Pierre-Emmanuel Mazare, and Jason Weston.
\newblock Learning from dialogue after deployment: Feed yourself, chatbot!
\newblock \emph{arXiv preprint arXiv:1901.05415}, 2019.

\bibitem[Hirth et~al.(2020)Hirth, Borchert, De~Moor, Borst, and
  Hoßfeld]{hirth2020taskprefs}
Matthias Hirth, Kathrin Borchert, Katrien De~Moor, Vanessa Borst, and Tobias
  Hoßfeld.
\newblock Personal task design preferences of crowdworkers.
\newblock In \emph{2020 Twelfth International Conference on Quality of
  Multimedia Experience (QoMEX)}, pp.\  1--6, 2020.
\newblock \doi{10.1109/QoMEX48832.2020.9123094}.

\bibitem[Hollender et~al.(2010)Hollender, Hofmann, Deneke, and
  Schmitz]{HOLLENDER2010hci}
Nina Hollender, Cristian Hofmann, Michael Deneke, and Bernhard Schmitz.
\newblock Integrating cognitive load theory and concepts of human–computer
  interaction.
\newblock \emph{Computers in Human Behavior}, 26\penalty0 (6):\penalty0
  1278--1288, 2010.
\newblock ISSN 0747-5632.
\newblock \doi{https://doi.org/10.1016/j.chb.2010.05.031}.
\newblock URL
  \url{https://www.sciencedirect.com/science/article/pii/S0747563210001718}.

\bibitem[Kiela et~al.(2021)Kiela, Bartolo, Nie, Kaushik, Geiger, Wu, Vidgen,
  Prasad, Singh, Ringshia, et~al.]{kiela2021dynabench}
Douwe Kiela, Max Bartolo, Yixin Nie, Divyansh Kaushik, Atticus Geiger,
  Zhengxuan Wu, Bertie Vidgen, Grusha Prasad, Amanpreet Singh, Pratik Ringshia,
  et~al.
\newblock Dynabench: Rethinking benchmarking in nlp.
\newblock \emph{arXiv preprint arXiv:2104.14337}, 2021.

\bibitem[Kojima et~al.(2021)Kojima, Suhr, and Artzi]{kojima2021continual}
Noriyuki Kojima, Alane Suhr, and Yoav Artzi.
\newblock Continual learning for grounded instruction generation by observing
  human following behavior.
\newblock \emph{Transactions of the Association for Computational Linguistics},
  9:\penalty0 1303--1319, 2021.

\bibitem[Mandlekar et~al.(2018)Mandlekar, Zhu, Garg, Booher, Spero, Tung, Gao,
  Emmons, Gupta, Orbay, et~al.]{mandlekar2018roboturk}
Ajay Mandlekar, Yuke Zhu, Animesh Garg, Jonathan Booher, Max Spero, Albert
  Tung, Julian Gao, John Emmons, Anchit Gupta, Emre Orbay, et~al.
\newblock Roboturk: A crowdsourcing platform for robotic skill learning through
  imitation.
\newblock In \emph{Conference on Robot Learning}, pp.\  879--893. PMLR, 2018.

\bibitem[Mandlekar et~al.(2020)Mandlekar, Xu, Mart{\'\i}n-Mart{\'\i}n, Zhu,
  Fei-Fei, and Savarese]{mandlekar2020human}
Ajay Mandlekar, Danfei Xu, Roberto Mart{\'\i}n-Mart{\'\i}n, Yuke Zhu,
  Li~Fei-Fei, and Silvio Savarese.
\newblock Human-in-the-loop imitation learning using remote teleoperation.
\newblock \emph{arXiv preprint arXiv:2012.06733}, 2020.

\bibitem[Mitchell et~al.(2018)Mitchell, Cohen, Hruschka, Talukdar, Yang,
  Betteridge, Carlson, Dalvi, Gardner, Kisiel, et~al.]{mitchell2018never}
Tom Mitchell, William Cohen, Estevam Hruschka, Partha Talukdar, Bishan Yang,
  Justin Betteridge, Andrew Carlson, Bhavana Dalvi, Matt Gardner, Bryan Kisiel,
  et~al.
\newblock Never-ending learning.
\newblock \emph{Communications of the ACM}, 61\penalty0 (5):\penalty0 103--115,
  2018.

\bibitem[Paxton et~al.(2017)Paxton, Hundt, Jonathan, Guerin, and
  Hager]{paxton2017costar}
Chris Paxton, Andrew Hundt, Felix Jonathan, Kelleher Guerin, and Gregory~D
  Hager.
\newblock Costar: Instructing collaborative robots with behavior trees and
  vision.
\newblock In \emph{2017 IEEE international conference on robotics and
  automation (ICRA)}, pp.\  564--571. IEEE, 2017.

\bibitem[Petrenko et~al.(2021)Petrenko, Wijmans, Shacklett, and
  Koltun]{petrenko2021megaverse}
Aleksei Petrenko, Erik Wijmans, Brennan Shacklett, and Vladlen Koltun.
\newblock Megaverse: Simulating embodied agents at one million experiences per
  second.
\newblock In \emph{International Conference on Machine Learning}, pp.\
  8556--8566. PMLR, 2021.

\bibitem[Pratik et~al.(2021)Pratik, Chintala, Srinet, Gandhi, Qian, Sun, Drew,
  Elkafrawy, Tiwari, Hart, et~al.]{pratik2021droidlet}
Anurag Pratik, Soumith Chintala, Kavya Srinet, Dhiraj Gandhi, Rebecca Qian,
  Yuxuan Sun, Ryan Drew, Sara Elkafrawy, Anoushka Tiwari, Tucker Hart, et~al.
\newblock droidlet: modular, heterogenous, multi-modal agents.
\newblock In \emph{2021 IEEE International Conference on Robotics and
  Automation (ICRA)}, pp.\  13716--13723. IEEE, 2021.

\bibitem[Radford et~al.(2021)Radford, Kim, Hallacy, Ramesh, Goh, Agarwal,
  Sastry, Askell, Mishkin, Clark, et~al.]{radford2021learning}
Alec Radford, Jong~Wook Kim, Chris Hallacy, Aditya Ramesh, Gabriel Goh,
  Sandhini Agarwal, Girish Sastry, Amanda Askell, Pamela Mishkin, Jack Clark,
  et~al.
\newblock Learning transferable visual models from natural language
  supervision.
\newblock In \emph{International Conference on Machine Learning}, pp.\
  8748--8763. PMLR, 2021.

\bibitem[Ratner et~al.(2017)Ratner, Bach, Ehrenberg, Fries, Wu, and
  R{\'e}]{ratner2017snorkel}
Alexander Ratner, Stephen~H Bach, Henry Ehrenberg, Jason Fries, Sen Wu, and
  Christopher R{\'e}.
\newblock Snorkel: Rapid training data creation with weak supervision.
\newblock In \emph{Proceedings of the VLDB Endowment. International Conference
  on Very Large Data Bases}, volume~11, pp.\  269. NIH Public Access, 2017.

\bibitem[Savva et~al.(2019)Savva, Kadian, Maksymets, Zhao, Wijmans, Jain,
  Straub, Liu, Koltun, Malik, et~al.]{savva2019habitat}
Manolis Savva, Abhishek Kadian, Oleksandr Maksymets, Yili Zhao, Erik Wijmans,
  Bhavana Jain, Julian Straub, Jia Liu, Vladlen Koltun, Jitendra Malik, et~al.
\newblock Habitat: A platform for embodied ai research.
\newblock In \emph{Proceedings of the IEEE/CVF International Conference on
  Computer Vision}, pp.\  9339--9347, 2019.

\bibitem[Saxena et~al.(2014)Saxena, Jain, Sener, Jami, Misra, and
  Koppula]{saxena2014robobrain}
Ashutosh Saxena, Ashesh Jain, Ozan Sener, Aditya Jami, Dipendra~K Misra, and
  Hema~S Koppula.
\newblock Robobrain: Large-scale knowledge engine for robots.
\newblock \emph{arXiv preprint arXiv:1412.0691}, 2014.

\bibitem[Shah et~al.(2021)Shah, Wild, Wang, Alex, Houghton, Guss, Mohanty,
  Kanervisto, Milani, Topin, et~al.]{shah2021minerl}
Rohin Shah, Cody Wild, Steven~H Wang, Neel Alex, Brandon Houghton, William
  Guss, Sharada Mohanty, Anssi Kanervisto, Stephanie Milani, Nicholay Topin,
  et~al.
\newblock The minerl basalt competition on learning from human feedback.
\newblock \emph{arXiv preprint arXiv:2107.01969}, 2021.

\bibitem[Shuster et~al.(2021)Shuster, Urbanek, Dinan, Szlam, and
  Weston]{shuster2021dialogue}
Kurt Shuster, Jack Urbanek, Emily Dinan, Arthur Szlam, and Jason Weston.
\newblock Dialogue in the wild: Learning from a deployed role-playing game with
  humans and bots.
\newblock In \emph{Findings of the Association for Computational Linguistics:
  ACL-IJCNLP 2021}, pp.\  611--624, 2021.

\bibitem[Simard et~al.(2017)Simard, Amershi, Chickering, Pelton, Ghorashi,
  Meek, Ramos, Suh, Verwey, Wang, et~al.]{simard2017machine}
Patrice~Y Simard, Saleema Amershi, David~M Chickering, Alicia~Edelman Pelton,
  Soroush Ghorashi, Christopher Meek, Gonzalo Ramos, Jina Suh, Johan Verwey,
  Mo~Wang, et~al.
\newblock Machine teaching: A new paradigm for building machine learning
  systems.
\newblock \emph{arXiv preprint arXiv:1707.06742}, 2017.

\bibitem[Srinet et~al.(2020)Srinet, Jernite, Gray, and
  Szlam]{srinet2020craftassist}
Kavya Srinet, Yacine Jernite, Jonathan Gray, and Arthur Szlam.
\newblock Craftassist instruction parsing: Semantic parsing for a voxel-world
  assistant.
\newblock In \emph{Proceedings of the 58th Annual Meeting of the Association
  for Computational Linguistics}, pp.\  4693--4714, 2020.

\bibitem[Suhr et~al.(2019)Suhr, Yan, Schluger, Yu, Khader, Mouallem, Zhang, and
  Artzi]{suhr2019executing}
Alane Suhr, Claudia Yan, Jacob Schluger, Stanley Yu, Hadi Khader, Marwa
  Mouallem, Iris Zhang, and Yoav Artzi.
\newblock Executing instructions in situated collaborative interactions.
\newblock \emph{arXiv preprint arXiv:1910.03655}, 2019.

\bibitem[Veniat et~al.(2020)Veniat, Denoyer, and Ranzato]{veniat2020efficient}
Tom Veniat, Ludovic Denoyer, and Marc'Aurelio Ranzato.
\newblock Efficient continual learning with modular networks and task-driven
  priors.
\newblock \emph{arXiv preprint arXiv:2012.12631}, 2020.

\bibitem[Wang et~al.(2017)Wang, Ginn, Liang, and Manning]{wang2017naturalizing}
Sida~I Wang, Samuel Ginn, Percy Liang, and Christoper~D Manning.
\newblock Naturalizing a programming language via interactive learning.
\newblock \emph{arXiv preprint arXiv:1704.06956}, 2017.

\bibitem[Wu et~al.(2021)Wu, Xiao, Sun, Zhang, Ma, and He]{hitl_survey}
Xingjiao Wu, Luwei Xiao, Yixuan Sun, Junhang Zhang, Tianlong Ma, and Liang He.
\newblock A survey of human-in-the-loop for machine learning.
\newblock \emph{arXiv preprint arXiv:2108.00941}, 2021.

\end{thebibliography}
\bibliographystyle{collas2022_conference}

\appendix
\label{sec:supplemental}
\clearpage

\section{Agent's Domain Specific Language(DSL)}
\label{sec:dsl}

This section describes the details of logical form of each action pictorially.
We support three dialogue types: HUMAN\_GIVE\_COMMAND, GET\_MEMORY and PUT\_MEMORY.
We support the following actions in our dataset : Build, Dance, Get, Spawn, Resume, Fill, Destroy, Move, Undo, Stop, Dig and FreeBuild.

In figure \ref{fig:event_fig}, we represent an event in the agent's grammar and DSL:
\begin{figure}[ht]
    \centering
    \includegraphics[width=16cm]{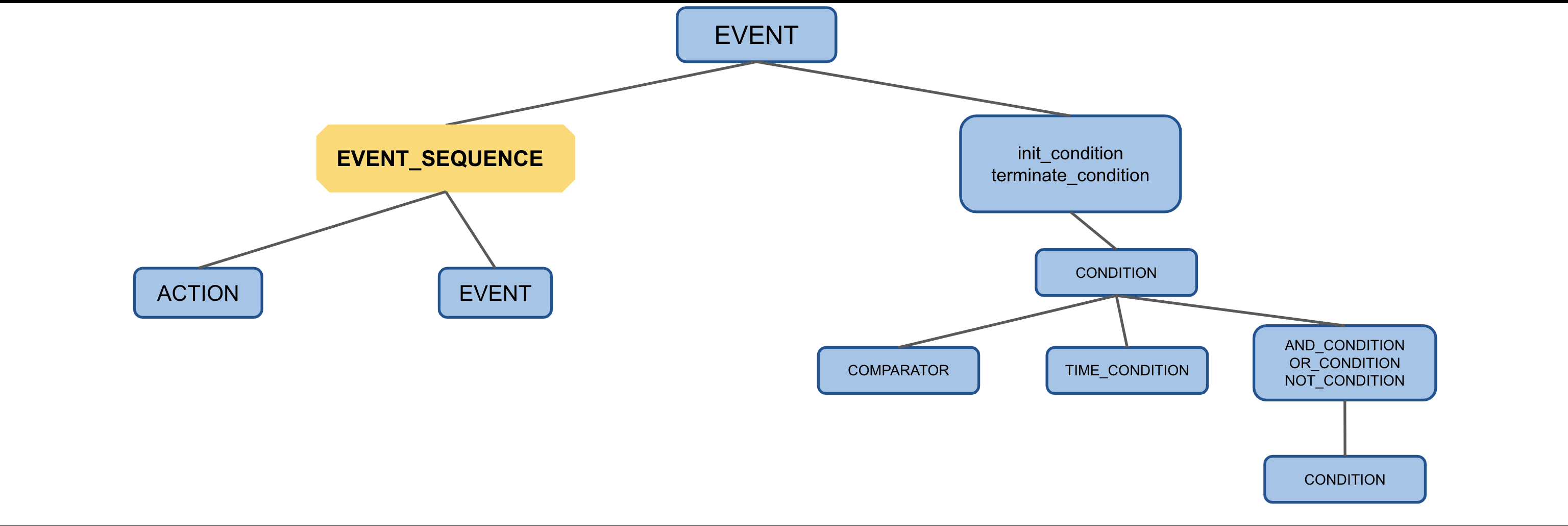}
    \caption{The agent's DSL showing the structure of an event.}
    \label{fig:event_fig}
\end{figure}

In figure \ref{fig:action_fig}, we show a full pictorial representation of actions in the agent's DSL:
\begin{figure}
    \centering
    \includegraphics[width=16cm]{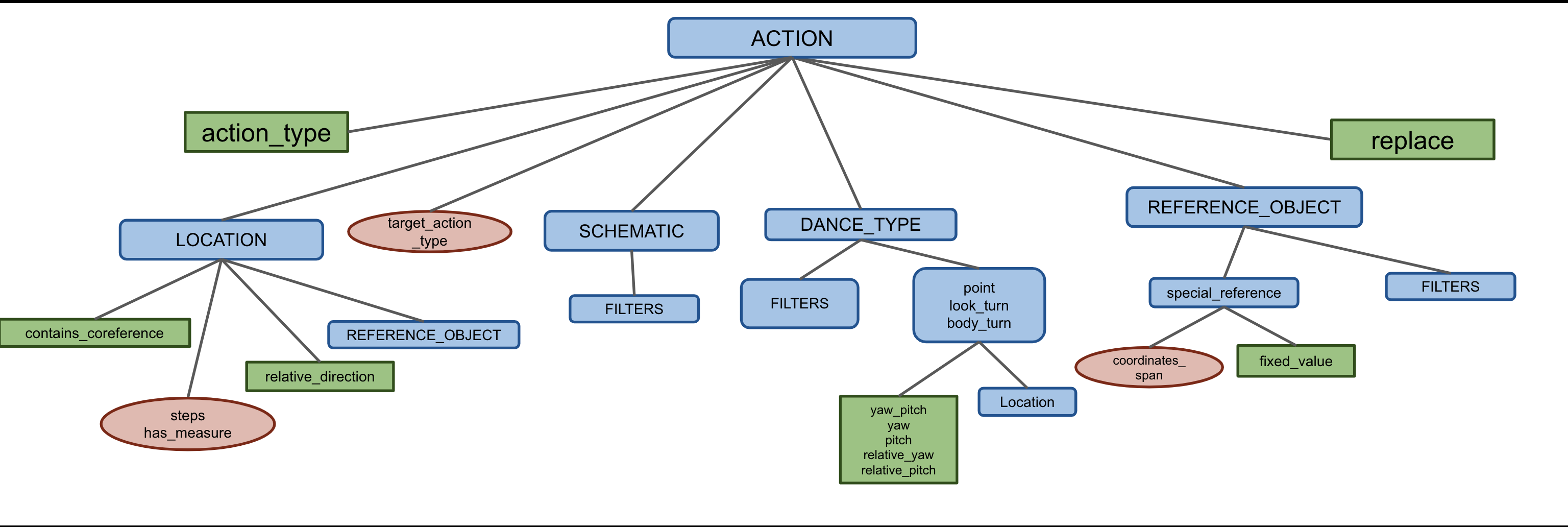}
    \caption{The representation of an action in agent's DSL.}
    \label{fig:action_fig}
\end{figure}

\begin{figure}
    \centering
    \includegraphics[width=16cm]{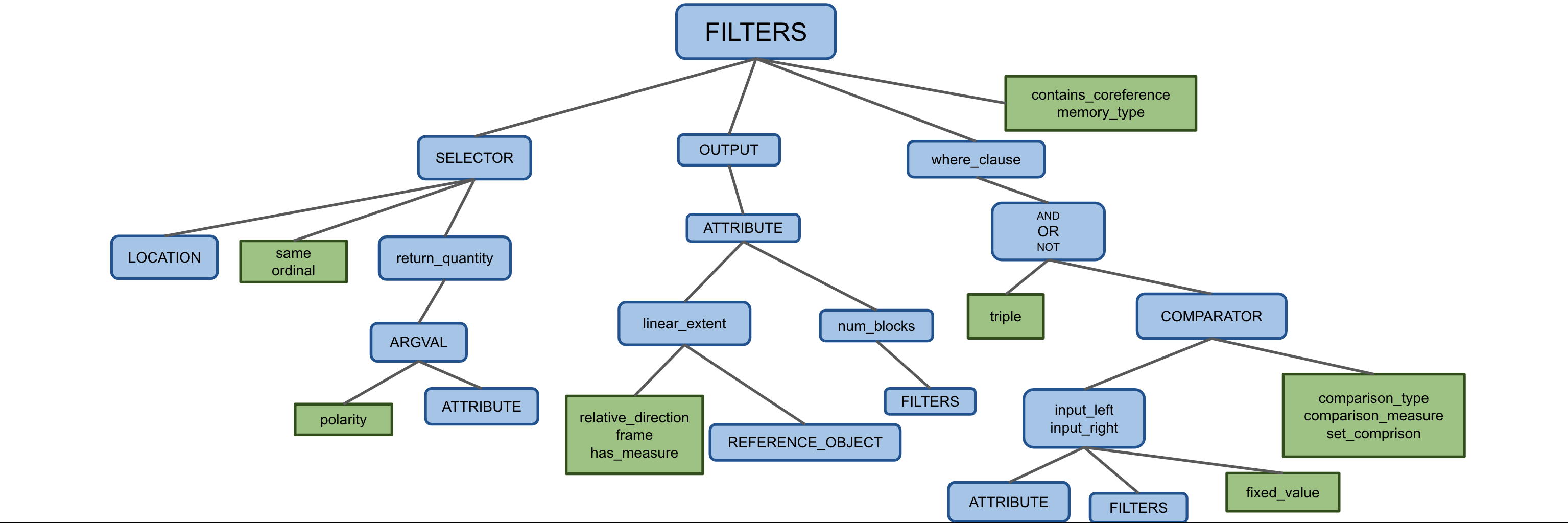}
    \caption{The representation of filters in agent's DSL.}
    \label{fig:filters_fig}
\end{figure}

Filters add a lot of expressiveness to the agent's grammar, we show a representation of filters in figure \ref{fig:filters_fig}

\clearpage

\section{Vision Module Details}
\label{sec:vision_module_detail}

\subsection{Vision Model Architecture}
This section describes the details of vision model architecture. The vision model consists of a Voxel Encoder to extract visual features and a Text Encoder to extract text features.

\textbf{Voxel Encoder} is 4-layer 3D convolutional neural network trained from scratch. For each layer, it consists of a 3D convolutional module with $kernel\_size=5$, $hidden\_dim=128$ and $padding=2$, followed by a 3D Batch Normalization module and a ReLU activation function to extract visual features $F_V\in\mathbb{R^{H \times W \times L \times C}} $ ($H$ denotes height, $W$ denotes width, $L$ denotes length and $C$ denotes hidden dimension). 

\textbf{Text Encoder} For any text input $ T\in\mathbb{R^L} $, we directly use a frozen CLIP \cite{radford2021learning} Text Encoder to extract text features  $ F_T\in\mathbb{R^{C \times L}} $, where $c=768$. Then we use a projection layer which transforms the $hidden\_dim$ from 768 to 128 to match the hidden dimension of Voxel Encoder output.  

Once we got both visual features $ F_V\in\mathbb{R^{H \times W \times L \times C}} $  and text features $ F_T\in\mathbb{R^{C \times L}} $, we perform matrix-matrix product on them followed by a $sigmoid$ layer to generate voxel-wise probability distribution $ P\in\mathbb{R^{H \times W \times L}} $ on how likely each voxel belongs to the object referred in the text. All voxels with probability high than $threshold=0.8$ are then used to construct the segmentation mask as model output.

\subsection{Vision Data Bootstrapping}

This section describes how we use rule-based scripts to generate initial training data which we use to train vision baseline models.

\begin{figure}
\centering
  \includegraphics[width=82mm]{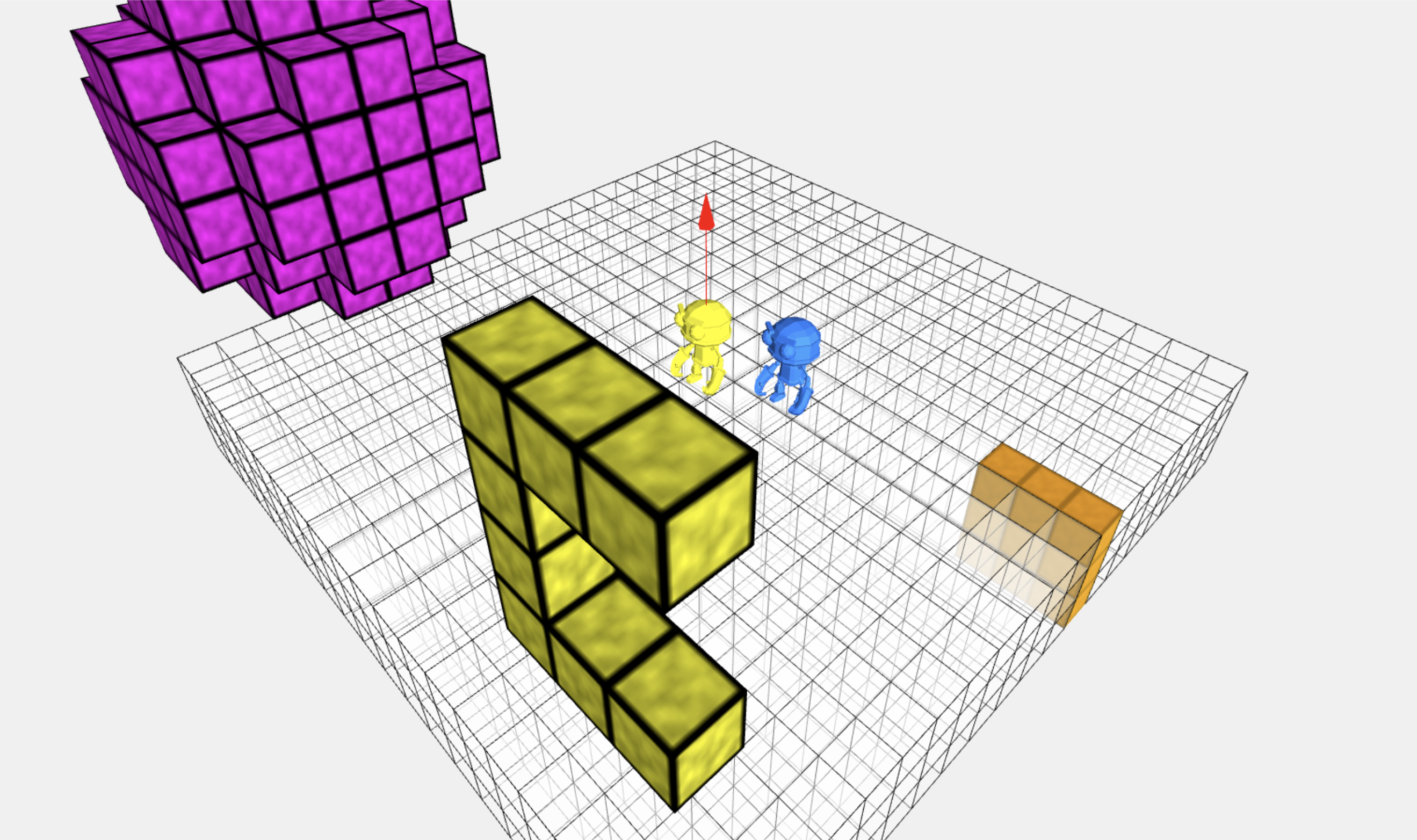}
  \includegraphics[width=82mm]{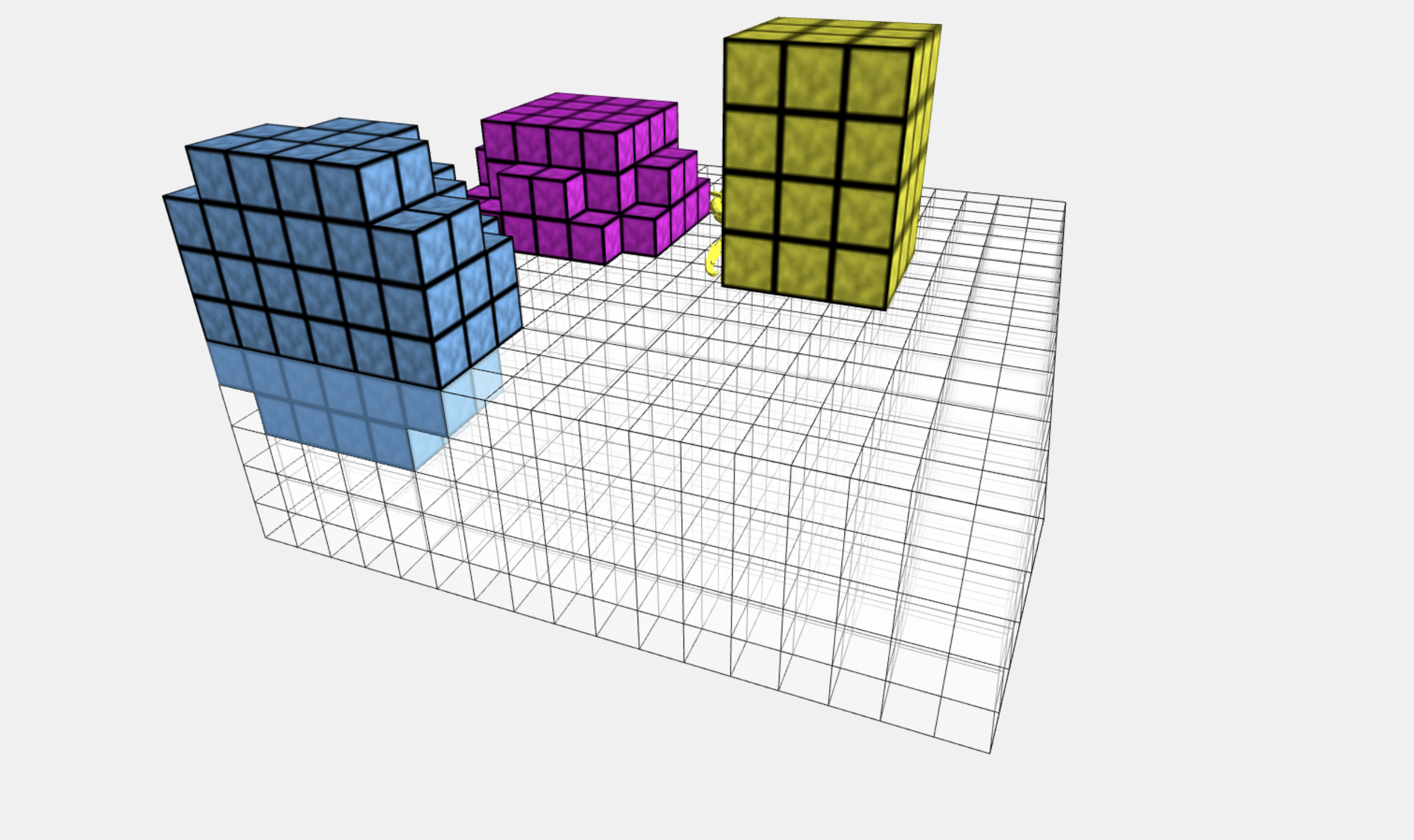}
\caption{Examples of templated vision data generated using rule-based scripts, rendered in our 3D visualization tool}
\label{fig:vision_data_example}
\end{figure}

We randomly put 1 to 3 shape objects with different materials and sizes into the 3D flat voxel world \ref{sec:world} and record the world state (i.e. voxel IDs at each voxel position) as schematics. The shape object is created purely based on mathematical formulas. For example, the schematic of a solid sphere with center $(x, y, z)$ and radius ${r}$ consists of all the points $(ix, iy, iz)$ in the space, where $(\sqrt{(ix - x) ^ 2 + (iy - y) ^ 2 + (iz - z) ^ 2}) <= r$. Figure \ref{fig:vision_data_example} shows  two examples of data points generated in this way, rendered in our 3D visualization tool.

The full list of shapes that we used is as follows: cube, rectanguloid, sphere, pyramid, square, rectangle, circle, triangle, dome, arch. We construct the text description of these shapes based on their names and colors (since it's rule based, we have prior knowledge of the material of each voxel and the color of it). For example, a cube made of gold blocks could be named as 'cube', 'yellow cube' or 'the yellow thing' randomly.

Furthermore, we construct some negative data points based on the positive ones, that is, we randomly ask objects that are not in the generated scenes. The segmentation masks of those non-existing objects are all empty.

\clearpage

\section{Data Generation Interface - Human Intelligence Task (HIT)}
\label{sec:hit}

This appendix describes the interface used by crowd-sourced workers to interact with the Droidlet agent and generate data by issuing commands.

\begin{figure}
    \centering
    \includegraphics[width=16cm]{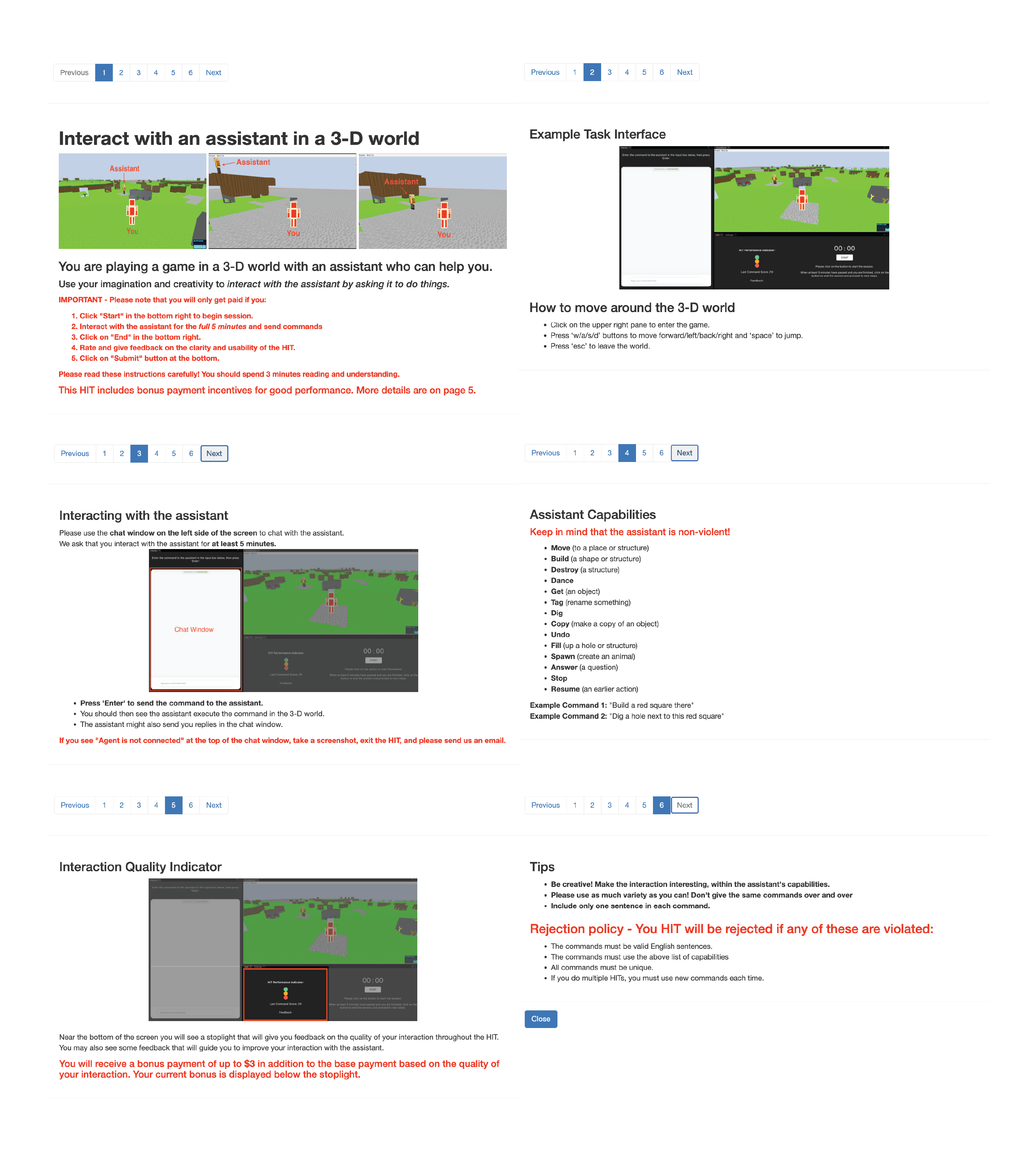}
    \caption{The instructions given to crowd-sourced workers for completing the agent interaction HIT.}
    \label{fig:instructions}
\end{figure}

Figure \ref{fig:instructions} is the instructions popup, which is the first thing that the worker sees when starting the HIT.  The instructions are paginated to reduce each section to a digestible amount of content.

\begin{figure}
    \centering
    \includegraphics[width=11cm]{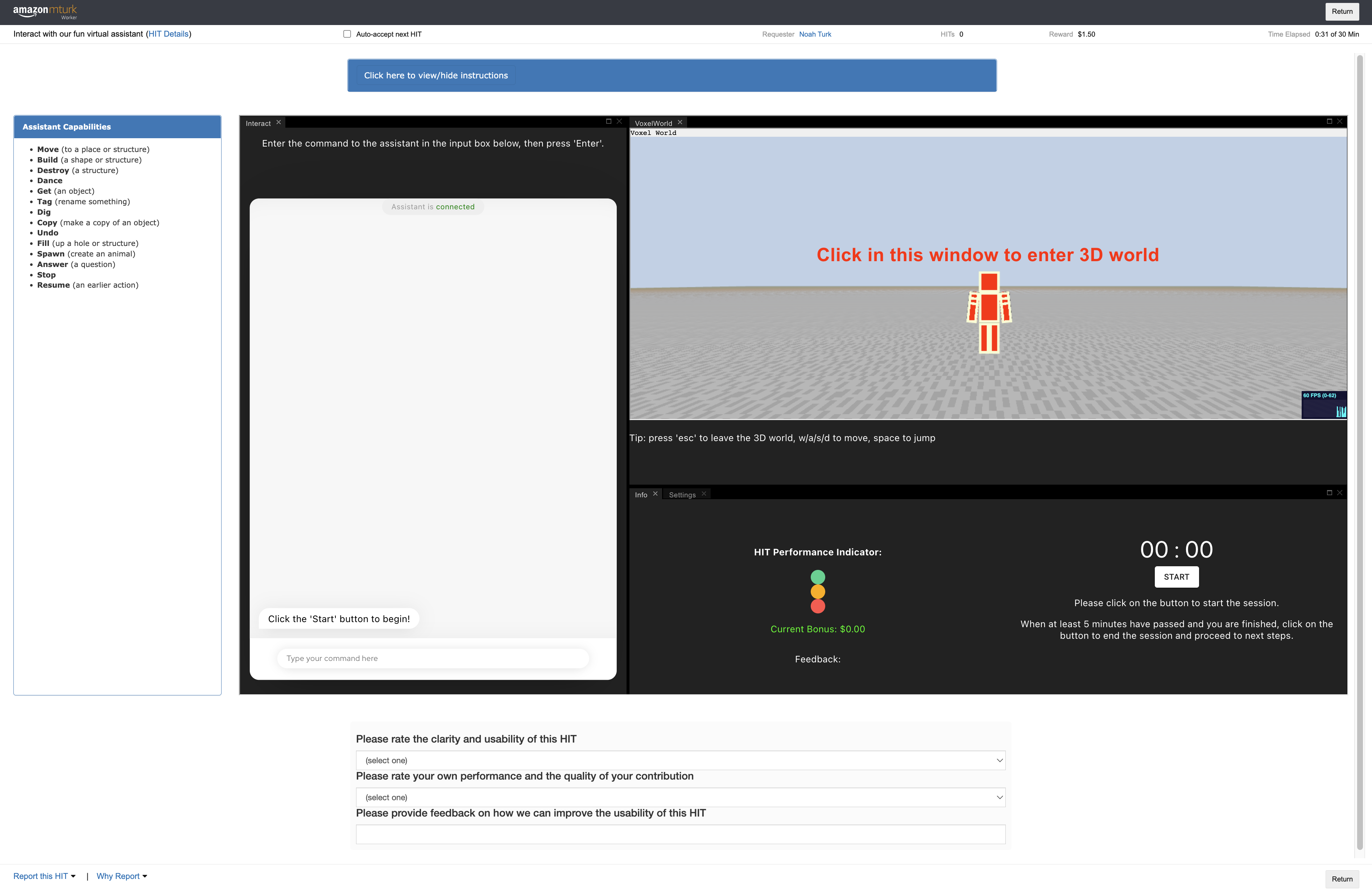}
    \caption{View of the HIT page at the beginning of the task.}
    \label{fig:start_hit}
\end{figure}

Figure \ref{fig:start_hit} shows the view of the HIT page that workers see at the beginning of the task.  There is a prompt in the chat to start the clock (each interaction is a minimum of five minutes), and there is a prompt superimposed over the voxel world window indicating that users need to click once in that window in order for the voxel world to render.  The "stoplight" performance score is a 0 out of 10 at the start of the HIT, and no feedback is available yet.  The instructions, which were shown in a popup previously, are available for review in the dropdown at the top of the page.  However, the agent capabilities are always available for easy reference just to the left of the interaction window.

\begin{figure}
    \centering
    \includegraphics[width=11cm]{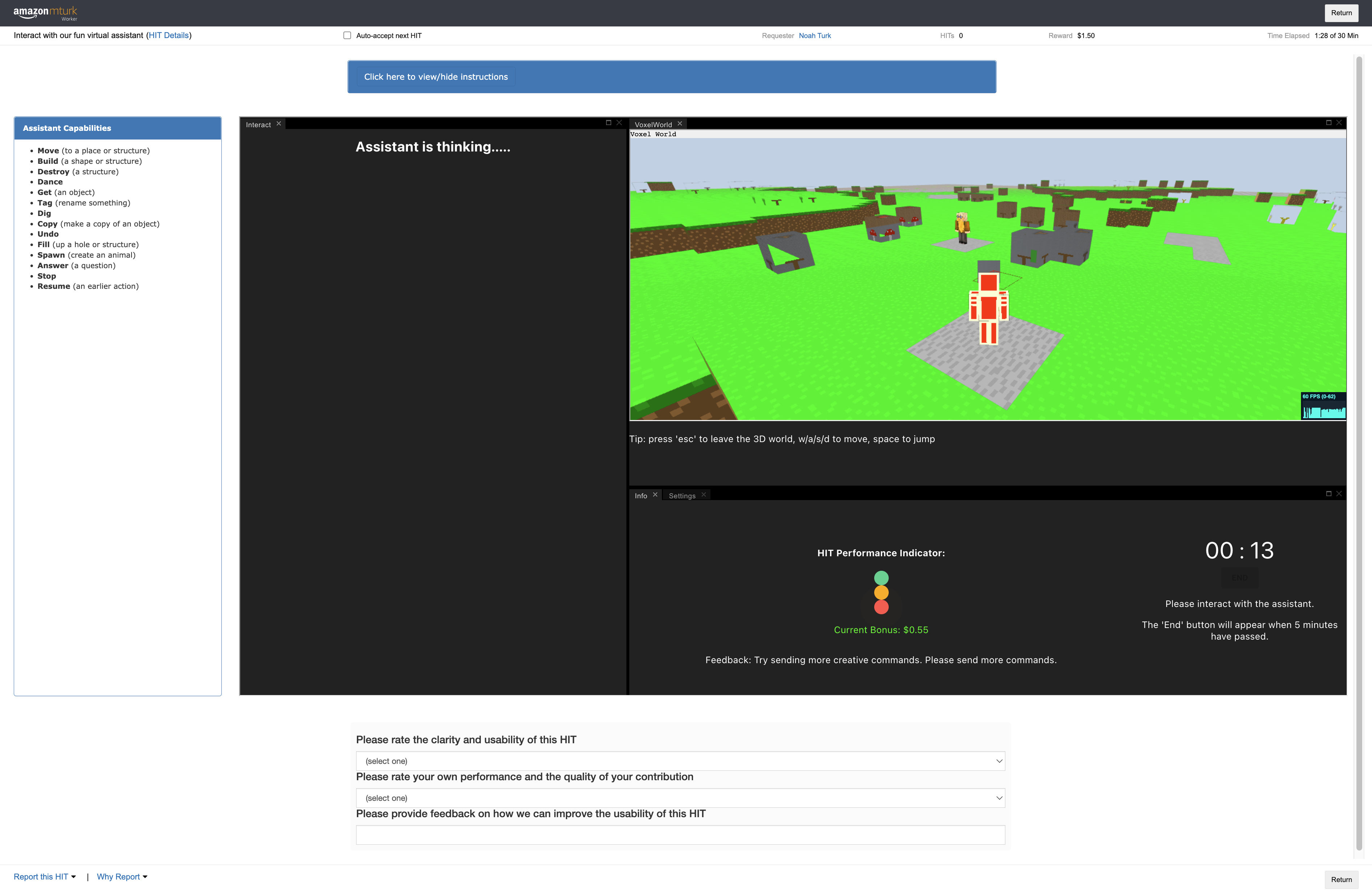}
    \includegraphics[width=11cm]{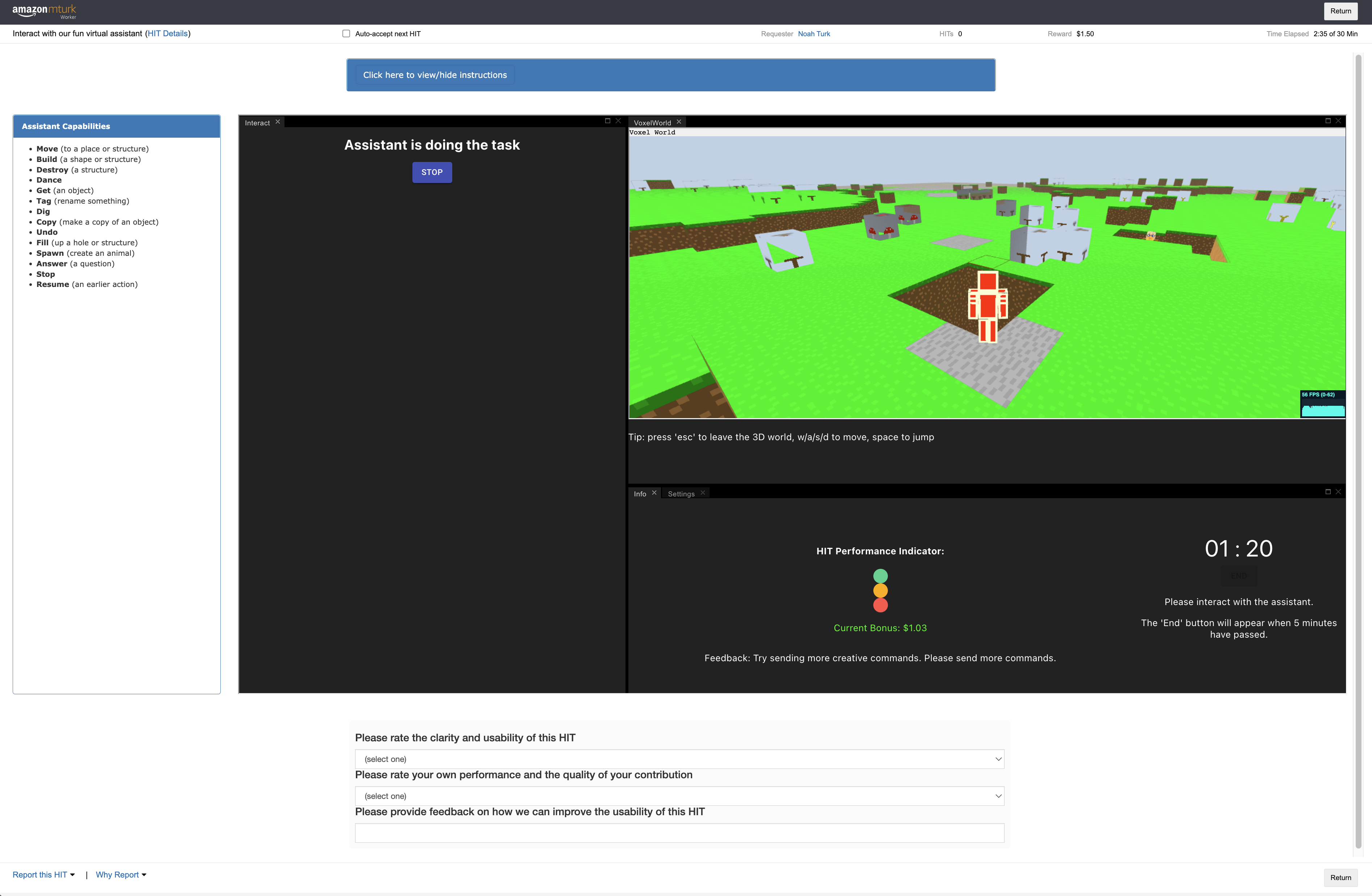}
    \caption{Status update messages given to workers as the agent is processing instructions and performing the task.  The worker retains the ability to issue a "stop" command while the agent is working.}
    \label{fig:status}
\end{figure}

Figure \ref{fig:status} shows two of the status messages that workers see after submitting a command.  These status messages are available so that the worker is not confused about what is happening at any given time, and can more reliably identify if there is a bug or the agent has frozen.  The four status messages that are shown after every command are, in order: "sending command", "command received", "assistant thinking", and "assistant is doing the task".  The first is cleared when the agent acknowledges having received the command.  The second is cleared after 500ms.  The third is cleared after the NSP has parsed the command.  The fourth and final status is cleared after the agent has completed the task, if it knows how.  After the fourth status message is cleared the next UI screen that appears is the error routing screen, which the user must progress through before being allowed to issue another command.

\begin{figure}
    \centering
    \includegraphics[width=11cm]{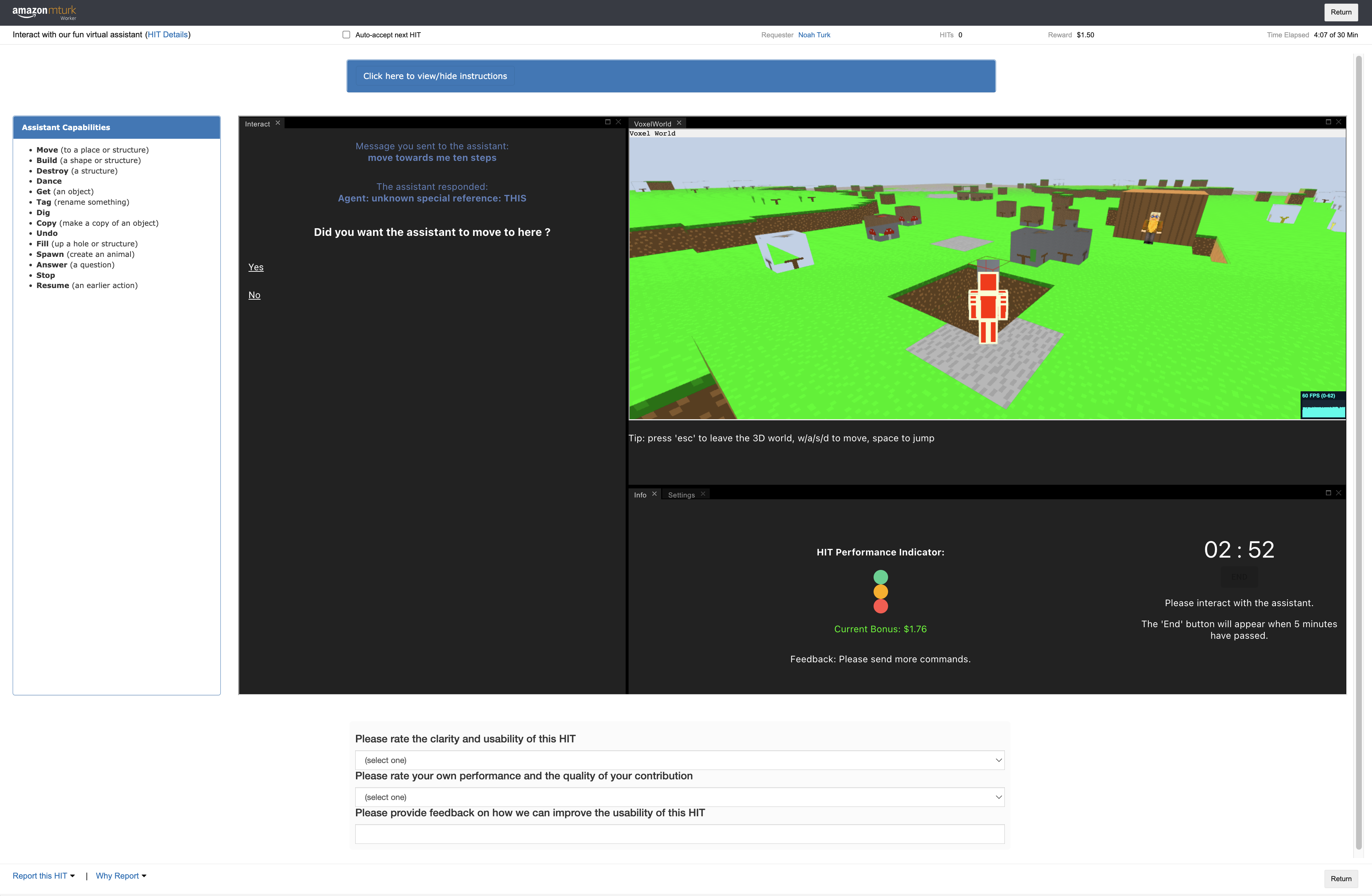}
    \includegraphics[width=11cm]{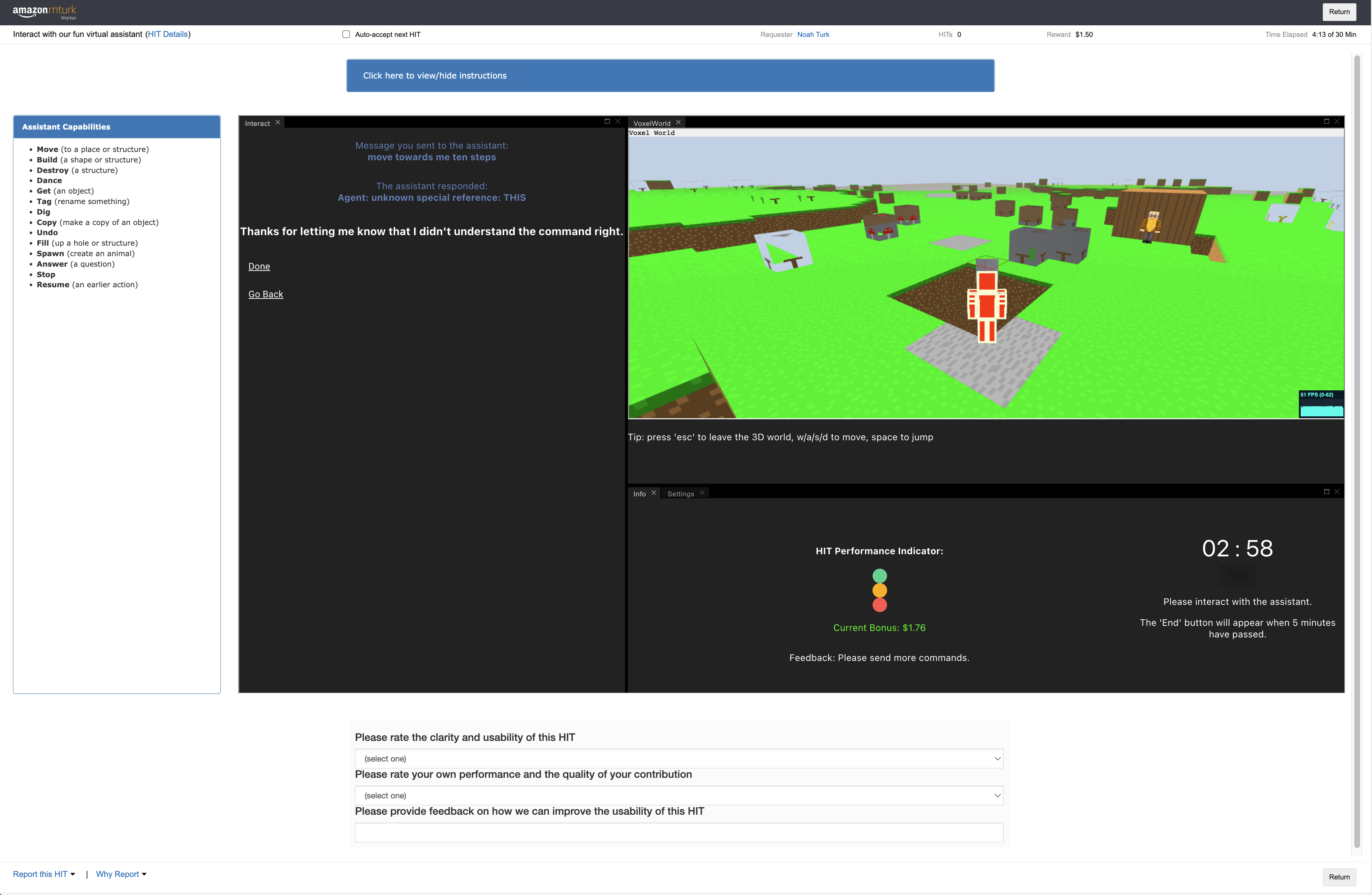}
    \caption{Error routing flow for a command that contains an NLU error.}
    \label{fig:nlu_error}
\end{figure}

\begin{figure}
    \centering
    \includegraphics[width=11cm]{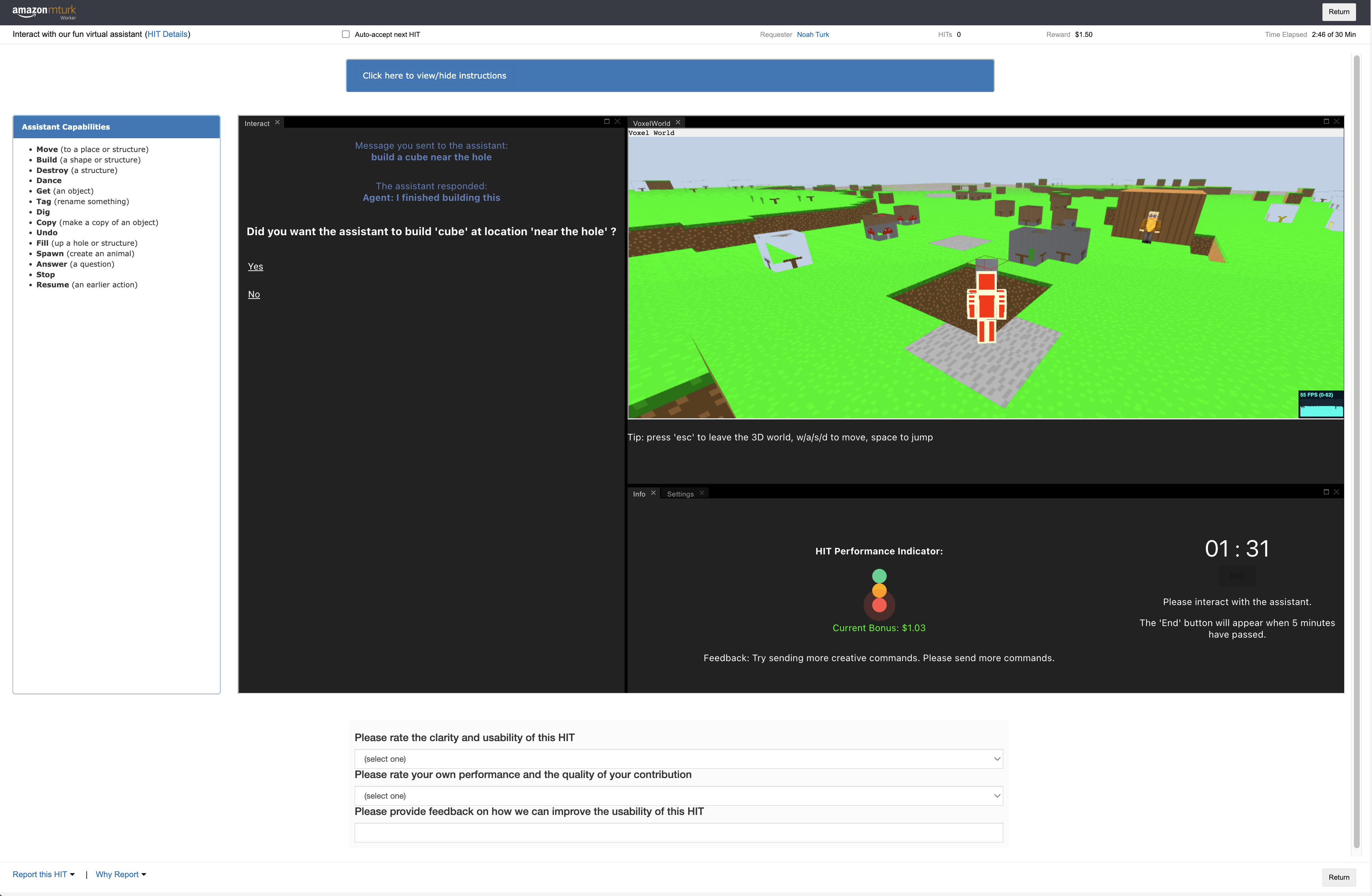}
    \includegraphics[width=11cm]{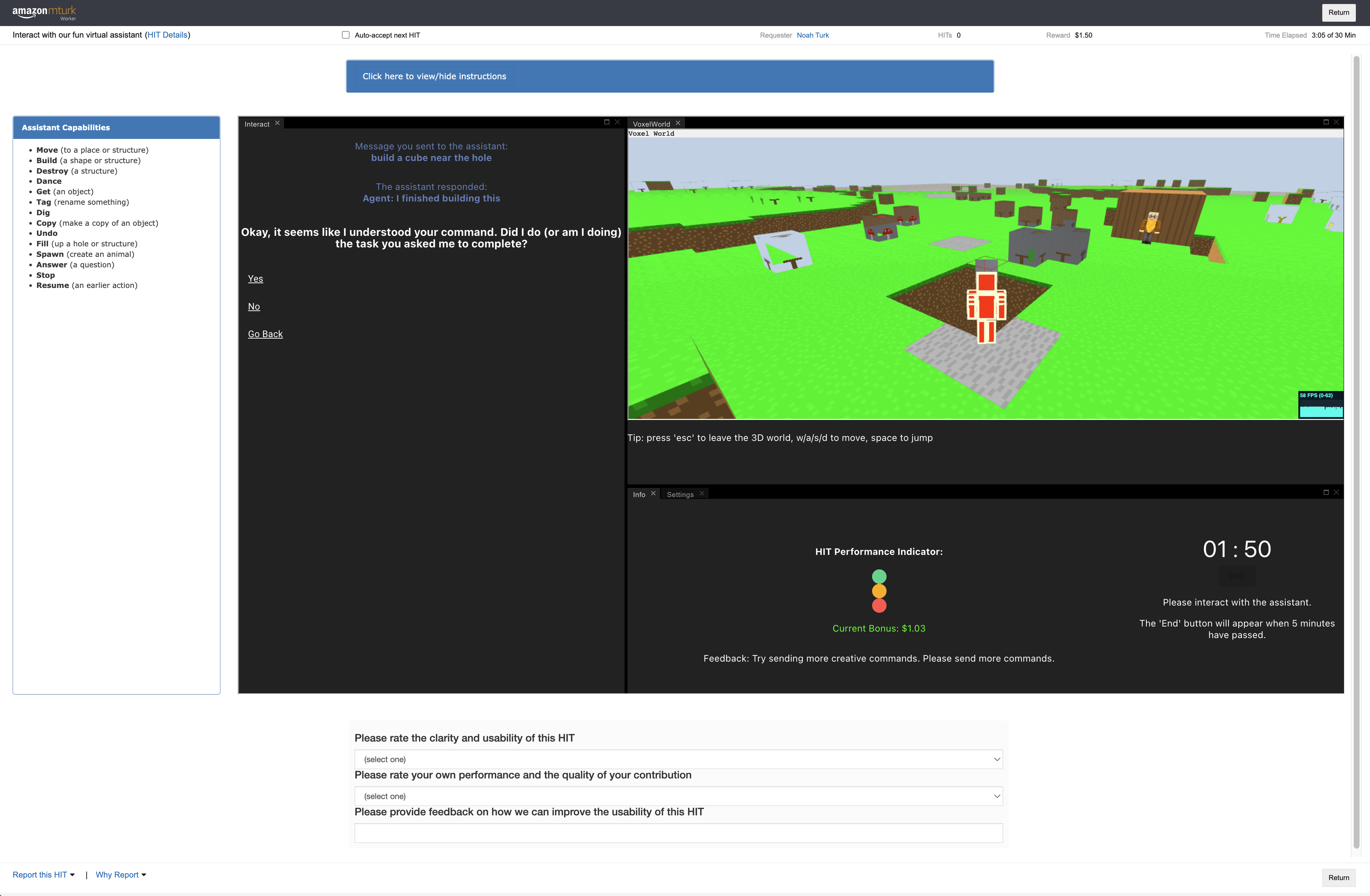}
    \includegraphics[width=11cm]{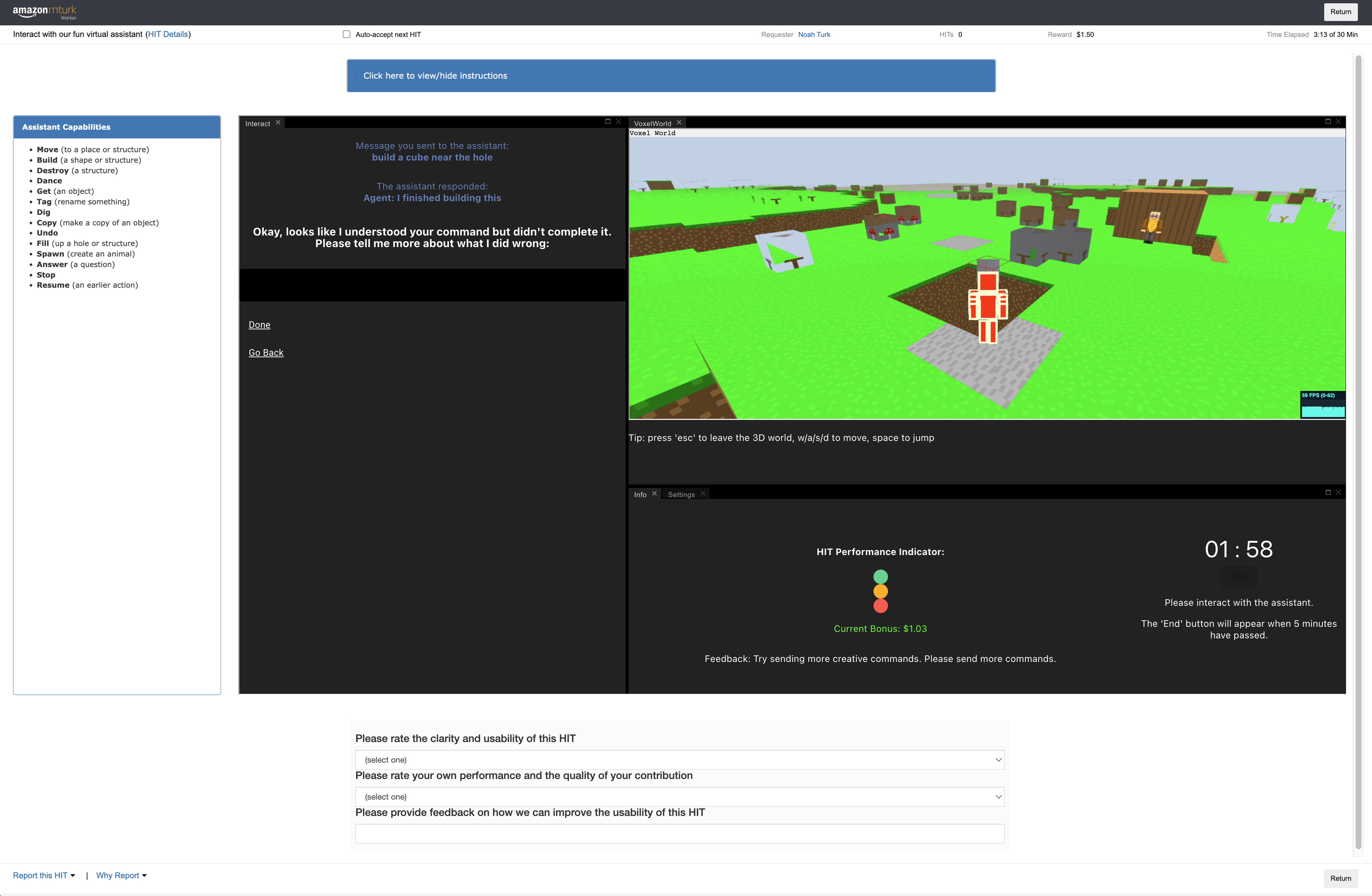}
    \caption{Error routing flow for a command that does not contain an NLU error but which the agent does not complete correctly (in this case a perception error).}
    \label{fig:task_error}
\end{figure}

Figure \ref{fig:nlu_error} and Figure \ref{fig:task_error} show the error marking flows after the agent processes a command containing an NLU error and a non-NLU task error, respectively.  Correct error marking is critical in order to appropriately route the appropriate data to the appropriate annotator.  After completing this decision tree presented one question at a time, the worker is returned back to the original interaction window shown in Figure 1.

\clearpage

\section{Vision Annotation Tool Details}
\label{sec:vision_anno_tool_detail}

\begin{figure}[ht]
    \centering
    \includegraphics[width=14cm]{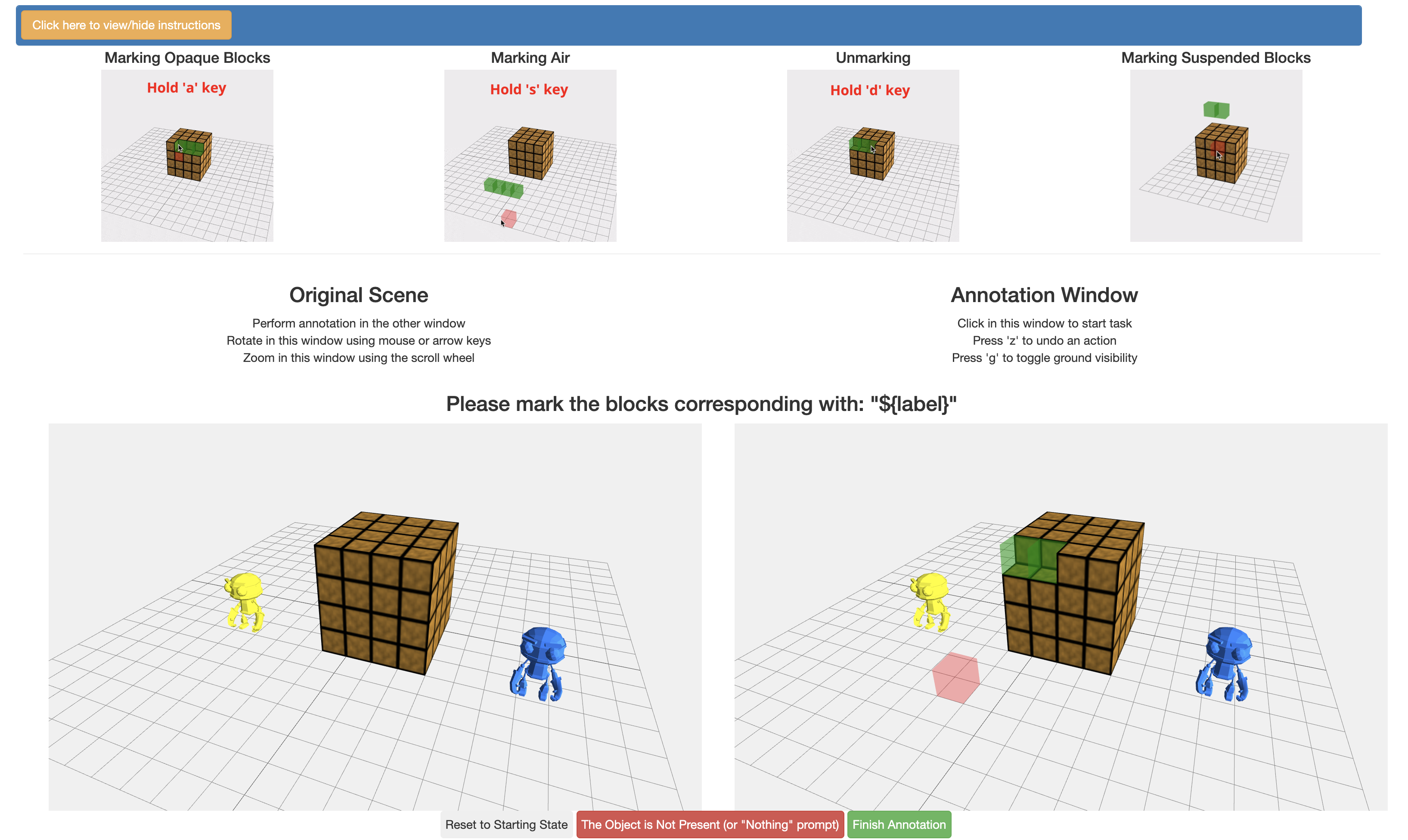}
    \caption{Vision annotation tool}
    \label{fig:vision_anno_tool_main}
\end{figure}

This section describes the details of vision annotation tool which crowdsourced workers used to annotate vision errors. For each object our vision model failed to recognize in a given scene, we extract the text description of that object and a snapshot of the 3D voxel world state and rerender in this web-based vision annotation tool. Crowdsourced workers are then asked to mark the blocks corresponding with the object as shown in figure \ref{fig:vision_anno_tool_main}.

\begin{figure}[ht]
    \centering
    \includegraphics[width=14cm]{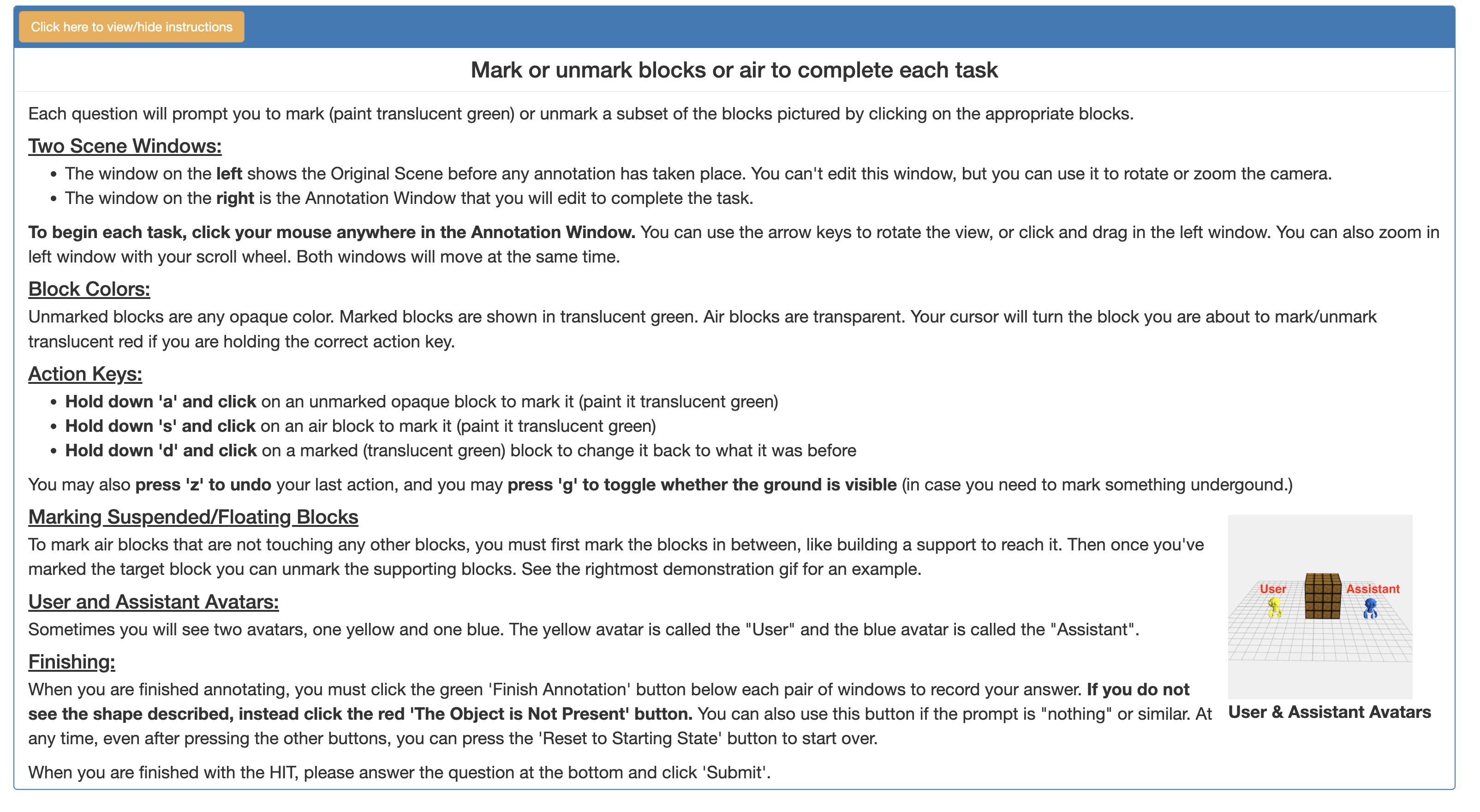}
    \caption{Vision annotation tool instructions}
    \label{fig:vision_anno_tool_instruction}
\end{figure}

Full instructions of how to use this tool are shown in figure \ref{fig:vision_anno_tool_instruction}.

\clearpage

\section{Model Training Details}
\label{sec:model_training_detail}
\subsection{NLU Module}
For the standard model retraining job, we train the models for 100 epochs (on average till lack of improvement on $V_n$). The batch size is set to 24 in order to fit into a single 16G GPU chip.

For Transformer decoder learning rate, we are choosing between 0.0000005, 0.000001 and 0.000005 while for the encoder learning rate we are choosing between 0.0 and 0.000001. This gives us a total number of 6 different combinations of hyperparameters for each model training job; we validate on $V_n$.  All other parameters are the default from \cite{huggingFace}. 

For model re-biasing, we train the models on the original training dataset for 10 epochs (this is roughly till lack of improvement on $V_0$, the initial validation set from \cite{srinet2020craftassist}).

\subsection{Vision Module}
For the standard model retraining job, we train the models for 4000 epochs (on average till lack of improvement on $V_n$). For model re-biasing, we train the models on the original training dataset for 50 epochs (this is roughly till lack of improvement on $V_0$, the initial validation set we generated using rule-based python scripts).

For hyperparameter search, we are choosing between 128 and 256 for voxel encoder hidden dimension size. For learning rate we are choosing between 0.001 and 0.0001. For probability threshold of being classified as positive we are choosing between 0.3, 0.5 and 0.8. This gives us a total number of 12 different combination. We trained the baseline vision models with those hyperparameters and this combination (hidden\textunderscore dim=128, learning\textunderscore rate=0.001, probability\textunderscore threshold=0.8) outperformed other combinations greatly, so we use this specific hyperparameter set for all the training jobs in the following iterations.

\end{document}